\documentclass[pdflatex,sn-mathphys-num]{sn-jnl}% M
\usepackage{adjustbox}
\usepackage{graphicx}%
\usepackage{multirow}%
\usepackage{amsmath,amssymb,amsfonts}%
\usepackage{amsthm}%
\usepackage{mathrsfs}%
\usepackage[title]{appendix}%
\usepackage{xcolor}%
\usepackage{textcomp}%
\usepackage{manyfoot}%
\usepackage{booktabs}%
\usepackage{caption}
\usepackage{algorithm}%
\usepackage{changepage}
\usepackage{float}
\usepackage{algorithmicx}%
\usepackage{algpseudocode}%
\usepackage{listings}%
\theoremstyle{thmstyleone}%
\theoremstyle{thmstyletwo}%

\theoremstyle{thmstylethree}%

\begin{document}

\title[Graph-Embedded Intuitionistic Fuzzy Broad Learning System: A Multi-view Framework]{Graph-Embedded Intuitionistic Fuzzy Broad Learning System: A Multi-view Framework}


\author[]{\fnm{Yogesh} \sur{Kumar}}\email{yogesh.23csz0014@iitrpr.ac.in}

\author[]{\fnm{Manju} \sur{}}\email{manju.24csz0020@iitrpr.ac.in}

\author*[]{\fnm{Mudasir} \sur{Ganaie}}\email{mudasir@iitrpr.ac.in}

\affil[]{\orgdiv{Department of Computer Science and Engineering}, \orgname{Indian Institute of Technology Ropar}, \orgaddress{\city{Rupnagar}, \postcode{140001}, \state{Punjab}, \country{India}}}

% \affil[2]{\orgdiv{Department}, \orgname{Organization}, \orgaddress{\street{Street}, \city{City}, \postcode{10587}, \state{State}, \country{Country}}}

% \affil[3]{\orgdiv{Department}, \orgname{Organization}, \orgaddress{\street{Street}, \city{City}, \postcode{610101}, \state{State}, \country{Country}}}


\abstract{The Broad Learning System (BLS) has been widely used for data classification and is based on a layer-by-layer feed-forward structure. However, it gives the same importance to all data points, which reduces its effectiveness on real-world datasets with noise and outliers. In addition, it does not consider the geometric structure of the data and has limitations in handling data from multiple sources. To address these challenges, we propose a Multi-View Graph-Embedded Intuitionistic Fuzzy Broad Learning System (MVGIFBLS) that integrates multi-view learning, graph embedding, and intuitionistic fuzzy theory into the BLS framework. This design enables the model to combine information from multiple sources and learn more discriminative representations. Graph embedding captures the geometric relationships among samples and improves class separation through intrinsic and penalty subspaces based on local Fisher discriminant analysis. Intuitionistic fuzzy theory enhances robustness to noise, while kernel-based neighborhood analysis captures local data structures. We evaluate the proposed framework on several UCI, KEEL, and AwA benchmark datasets using comparative evaluation, Gaussian feature noise analysis, ablation studies, and statistical analysis. The results demonstrate that each component contributes positively to the overall framework and that the proposed MVGIFBLS consistently achieves higher Area Under the Curve (AUC) scores and maintains robust performance under Gaussian feature noise.}


\maketitle
\section{Introduction}
Artificial neural networks (ANNs) are learning models based on how the human brain works. They use interconnected nodes, called neurons, to find patterns in the data and make predictions for tasks like recognizing images or understanding speech. Multi-layer ANNs use multiple layers to solve complex problems such as natural language processing \cite{Devlin2019BERTPO}, speech recognition \cite{adolfi2023successes}, and feature interpretation \cite{10416391}. It extracts meaningful features from raw data with minimal human effort and achieves accurate predictions on unseen data \cite{gallicchio2020deep}.However, ANNs have significant challenges, as they often contain millions of parameters and many layers, making them complex to design and train.Training relies on iterative methods such as gradient descent, which may be slow and computationally expensive. It also requires high-performance hardware, such as GPUs or TPUs, which may not always be available.These challenges limit the effectiveness of ANNs and highlight the need for simpler and more efficient methods.

Randomized neural networks (RNNs) address these issues by randomly assigning some weights and biases and keeping them fixed during training. This reduces computational cost and speeds up the learning process \cite{cao2018review,suganthan2021origins}. Two well-known RNN models are the Extreme Learning Machine (ELM) \cite{huang2006extreme,wang2022review} and the Random Vector Functional Link (RVFL) network \cite{malik2023random,subramani2025improving}. ELM computes the output weights in a single step using a closed-form solution \cite{wang2022review,huang2006extreme}. The RVFL network improves this approach by adding direct connections from the input to the output layer, which helps reduce overfitting and improve accuracy \cite{zhang2016comprehensive}. The RVFL network is effective due to its simple structure, fast training, and ability to approximate any function \cite{igelnik1995stochastic}.

Inspired by the RVFL network, the Broad Learning System (BLS) \cite{chen2017broad} is developed as a horizontally expanded neural network that processes data layer by layer. BLS has three core components: feature learning, enhancement, and output layers. In feature learning, the model transforms the input data into a higher-dimensional space using random projections. The enhancement layer applies nonlinear transformations to capture complex patterns. Finally, the output layer weights are optimized using the least-squares method, without using backpropagation. This design reduces overfitting, accelerates training, and enables BLS to perform well with different datasets. BLS supports incremental updates, enabling it to incorporate new data without retraining the full model. Its universal approximation capability ensures consistent performance in various applications \cite{chen2018universal}. Many improved versions of BLS have been proposed to solve different problems, making it more flexible for use in various applications. A double-kernelized BLS improves performance on imbalanced datasets by enhancing feature extraction, which is useful when some classes have fewer samples \cite{saputra2024blsf}. A multimodal BLS can process different types of data, such as texture information, allowing robots to sense materials through touch. This is useful in advanced manufacturing and exploration tasks \cite{wang2021multi}. The time-varying BLS variants adjust their settings step by step to manage changing conditions, making them ideal for controlling complex systems such as self-driving cars \cite{feng2018broad}. Neurofuzzy BLS models use fuzzy logic with BLS to make decisions like humans, improving how clearly the system explains its predictions for tasks such as disease diagnosis \cite{feng2018fuzzy}. In addition, deep-randomized structures that improve BLS by adding deeper layers, helping to learn more detailed patterns for complex datasets \cite{gallicchio2020deep}. A graph-based classification technique uses data connections to improve predictions, such as identifying objects in images \cite{liu2021graph}. Ensemble methods increase reliability by combining multiple BLS models, making predictions more accurate and stable \cite{chu2024efficient}. Session-incremental EEG classification adapts BLS for analyzing brain signals over time, supporting brain-computer interface applications \cite{yang2024session}. Stacking with residual structures adds extra layers to BLS to refine classification results and improve performance \cite{huang2024stacking}. Robust label noise handling reduces errors caused by mislabeled data \cite{deng2025robust}. Graph embedding preserves structural information in the data and helps handle imbalanced classes more effectively, which is useful in applications such as fault detection \cite{shi2023graph}. Local embedding with graph convolution optimizes BLS for hyperspectral image classification, enhancing satellite image analysis \cite{li2023local}. Fast graph embedding speeds up node classification in network data, improving efficiency in social network analysis \cite{jiang2019fast}. Although these updates make BLS more versatile, some methods, such as fuzzy clustering, can increase the time and resources needed, posing a challenge for large-scale use \cite{feng2018fuzzy}. Recent research has proposed several new BLS models. An ensemble BLS based on structural diversity reduces memory use when many models are stored and combined \cite{chu2024efficient}. Another model, called the BLS Learning Machine, uses an improved optimizer to remove hand tremors in teleoperation by reducing signal errors \cite{yang2021broad}. For chaotic time series, a multi-head attention BLS model helps the network learn time-based patterns and gives better predictions than LSTM and the basic BLS \cite{su2023multi}. BLS has also been used to predict cement setting time from material properties, helping to reduce manual work in construction \cite{guo2020efficient}. In addition, a BLS model with a Takagi–Sugeno fuzzy subsystem has been used to identify tobacco origins from near-infrared data, and can complete training quickly while giving useful results for quality control \cite{wang2023broad}.

BLS offers several advantages over traditional machine learning and deep learning models \cite{gong2021research}. However, it is sensitive to noise and outliers, where outliers are data points that differ from normal data, and noise refers to random errors or fluctuations in the data \cite{smiti2020critical}. BLS is sensitive to noise and outliers for two main reasons. First, noisy features propagate through the feature and enhancement nodes and mix with clean features, which degrades the learned representations. Second, BLS assigns equal importance to all training samples without distinguishing between clean and noisy data. As a result, the model struggles to separate useful information from noise, leading to reduced generalization performance.

Intuitionistic Fuzzy (IF) theory is widely used in machine learning to reduce the impact of noise and outliers on model performance \cite{sajid2024intuitionistic,zhang2023tsk}. It assigns a score to each sample using both membership and non-membership functions. The membership value shows how strongly a sample belongs to a class, while the non-membership value shows how strongly it does not belong. IF has been successfully applied in tasks such as Alzheimer’s diagnosis and classification under noisy conditions \citep{malik2022alzheimer,guo2024intuitionistic}, showing its effectiveness in handling noise and outliers.

In many real-world applications, data can be described using multiple feature sets, which leads to multiview data. For example, an image can be represented by color and texture features, a person can be identified using facial or fingerprint information, and a webpage can be described by its text content as well as the anchor text of incoming links \cite{xu2017re}. Although a single view may sometimes not provide satisfactory results, combining multiple views often leads to better performance by exploiting complementary information. To address this, multiview learning (MVL) emerged as a strong method to make use of data from multiple forms \cite{houthuys2018multi}. MVL methods follow two main principles: the consensus principle, which encourages agreement among different views to improve learning stability, and the complementarity principle, which uses the unique and complementary information from each view to improve prediction accuracy \cite{yu2025review}.

Many real-world datasets exhibit an intrinsic geometric structure, where preserving both local and global relationships among data points is important. BLS is efficient and scalable, but it often struggles to capture these patterns, which limits its ability to represent data effectively.  To address this limitation, integrating the graph embedding (GE) framework  \cite{yan2006graph} into the BLS increases
its learning abilities by embedding the intrinsic geometric structure of the data into the feature space. This combination allows BLS to preserve both local and global geometric relationships, which improves its representation and generalization performance. In addition, GE handles data variations and class separation, while intuitionistic fuzzy methods help manage noise and outliers \cite{chen2025adaptive,huang2023gfbls}. Another study uses GE to capture class structure and relationships, which improves feature extraction and shows strong performance on specific datasets \cite{shi2023graph}. Jiang et al. \cite{jiang2019fast} propose a GE-based BLS model that uses graph embedding to capture class structure and relationships for improved feature extraction. They show its effectiveness in specific datasets, such as data from rotating machinery.

Existing BLS methods, however, have complementary limitations. MVL-based approaches \cite{houthuys2018multi} combine information across views but often miss structural patterns and handle noise poorly. GE-based methods \cite{jiang2019fast} capture geometric relationships effectively but are less robust to noise and outliers. IF-based methods \cite{sajid2024intuitionistic} handle noise well but do not exploit multi-view information or structural patterns. Consequently, addressing only one of these limitations is insufficient for learning robust representations from complex multi-view data. To overcome this, we propose the Multi-View Graph-Embedded Intuitionistic Fuzzy Broad Learning System (MVGIFBLS), which incorporates MVL, GE, and IF theory within the BLS framework. The proposed framework jointly exploits complementary information from multiple views, preserves the intrinsic geometric structure of the data, and improves robustness to noise and outliers through adaptive sample weighting. This unified design enables the model to learn more discriminative feature representations and achieve improved classification performance.

The main contributions of this work are:
\begin{itemize}

  \item We propose MVGIFBLS, which integrates multi-view learning, graph embedding, and intuitionistic fuzzy theory within the Broad Learning System (BLS) framework.

    \item The proposed MVGIFBLS incorporates a multi-view learning strategy and a cross-view consistency term to improve classification performance.

    \item To preserve the intrinsic geometric structure of the data, graph embedding is incorporated into the framework. By utilizing intrinsic and penalty graphs with Local Fisher Discriminant Analysis (LFDA) \cite{sugiyama2007dimensionality}, the proposed framework preserves neighborhood relationships and improves class separation.

    \item Intuitionistic fuzzy theory is integrated into the framework to reduce the influence of noise and outliers through kernel-based neighborhood analysis for sample weighting.

    \item We evaluate the proposed MVGIFBLS on benchmark datasets from the UCI \cite{asuncion2007uci}, KEEL \cite{derrac2015keel}, and AwA \cite{lampert2013attribute} repositories. In addition, Gaussian feature noise experiments, ablation studies, and statistical analyses are conducted to evaluate robustness and analyze the contributions of graph embedding, intuitionistic fuzzy weighting, cross-view consistency, and multi-view feature fusion.
\end{itemize}

The layout of the paper is as follows: Section \ref{section2} presents a detailed overview of BLS, MVL, GE, and IF schemes. Section \ref{section3} provides the mathematical foundations of MVGIFBLS, LFDA with graph embedding, and the corresponding framework algorithm. Section \ref{section4} provides a detailed discussion of the results. Finally, Section \ref{section5} concludes the paper and suggests future research directions.

\section{Preliminary Work}
\label{section2}
% \begin{center}
% \textbf{RELATED WORKS}
% \end{center}
In this section, we first explain the notations used and then describe the BLS \cite{chen2017broad} in detail. We also provide an overview of MVL, discuss the GE framework \cite{yan2006graph}, and explain intuitionistic fuzzy \cite{guo2024intuitionistic} theory.

\subsection{Notations}
For better understanding, we define the following notation for a dataset with two views. Suppose we have a training dataset \( D \), written as \( D = \left\{ (v_i^E, v_i^F, t_i) \mid v_i^E \in \mathbb{R}^{1 \times m}, v_i^F \in \mathbb{R}^{1 \times n}, t_i \in \{-1, +1\}, i = 1, 2, \ldots, N \right\} \), 
where \( v_i^E \) and \( v_i^F \) are the feature vectors of the \( i^{th} \) data sample for view \( E \) and view \( F \), respectively. \( t_i \) is the label for that sample in both views. Let \( V_E \in \mathbb{R}^{N \times m} \) and \( V_F \in \mathbb{R}^{N \times n} \) be the input matrices for both views, where \( N \) is the total number of samples. The ground truth labels for all samples are written as \( T_{\text{true}} \in \{-1, +1\}^N \).

\subsection{Broad Learning System}

\begin{figure}[htbp]
    \centering
    \includegraphics[width=1.0\textwidth]{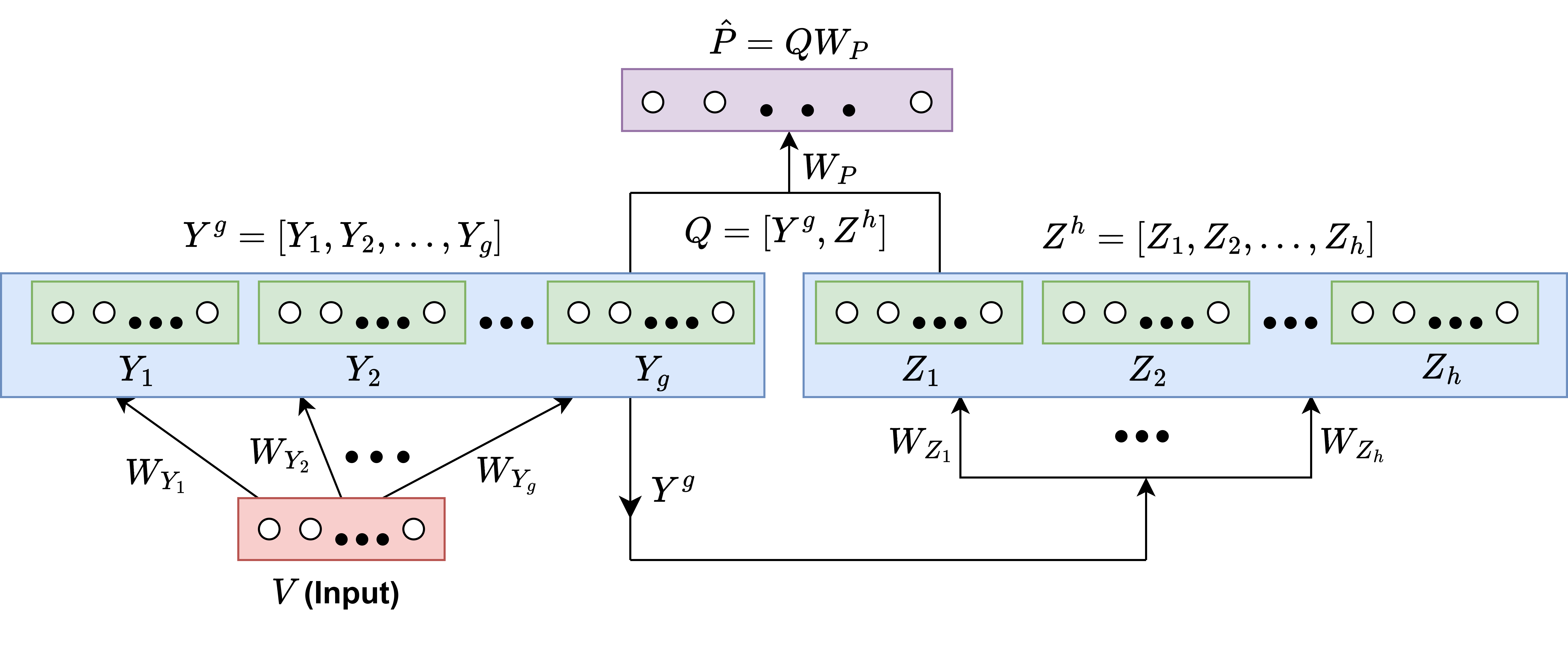}
    \caption{Block diagram of the BLS model}
    \label{BLS}
\end{figure}

The structure of the Broad Learning System (BLS) model \cite{chen2017broad} is shown in Figure \ref{BLS}, followed by a brief description of its mathematical formulation.

\textit{1)} Suppose there are $g$ feature groups, and each group has $r$ nodes. The $i$-th feature group can be written as:

\begin{equation}
Y_i = \phi_i( V W_{Y_i} +b_{Y_i}) \in \mathbb{R}^{N \times r}, \quad i = 1, 2, \dots, g,
\end{equation}
\noindent
here $\phi_i$,  $W_{Y_i} \in \mathbb{R}^{m \times r} $, and $b_{Y_i} \in \mathbb{R}^{N \times r}$ represent the feature map, the randomly initialized weight matrix, and the bias matrix for the $i$-th feature group, respectively. The augmented output of the $g$ number of feature groups is calculated as follows:
\begin{equation}
Y^g = [Y_1, Y_2, \dots, Y_g] \in \mathbb{R}^{N \times gr}.
\end{equation}
\textit{2)} The augmented feature matrix $Y^g$ is transformed into enhancement spaces using random transformations and activation functions. Let $h$ be the number of enhancement groups, and each group has $s$ nodes. The output of the $j$-th enhancement group is calculated as follows: 
\begin{equation}
    Z_j = \psi_j (Y^g W_{Z_j} + b_{Z_j}) \in \mathbb{R}^{N \times s}, \quad j = 1, 2, \ldots, h,
\end{equation}
where $\psi_j$ is the activation function, $W_{Z_j} \in \mathbb{R}^{gr \times s}$ is a randomly created weight matrix connecting $Y^g$ to the $j$-th enhancement group, and $b_{Z_j} \in \mathbb{R}^{N \times s}$ is the bias matrix for the same group. The combined output of all $h$ enhancement groups is calculated as follows:
\begin{equation}
Z^h = [Z_1, Z_2, \dots, Z_h] \in \mathbb{R}^{N \times hs}.
\end{equation}

\textit{3)} The enhancement matrix $Z^h$, along with the augmented feature matrix $Y^g$, are sent to the output layer. The final output is calculated as:
\begin{equation}
  \hat{P} = QW_P. 
\end{equation}

where
\begin{equation}
Q = [Y^g, Z^h] \in \mathbb{R}^{N \times (gr+hs)}    
\end{equation}
here, $Q$ is the concatenated matrix, and $W_P \in \mathbb{R}^{(gr+hs) \times C}$ is the weight matrix that links the augmented feature layer and the enhancement layer to the output layer. The weight matrix $W_P$ is found using the inverse of $Q$, given as $W_P = Q^\dagger \hat{P}$,
where $Q^\dagger$ represents the pseudoinverse of $Q$.

\subsection{Review on  multi-view learning}
Multi-view learning (MVL) is an active area of machine learning that improves model performance by combining data from different feature sources. By combining multiple views, MVL creates richer feature representations, which improve model accuracy and reliability \cite{zhao2017multi}. In practice, data samples often come from different feature sets or spaces, such as images described by color and texture, or webpages described by text and links \cite{li2016low}. Traditional approaches usually merge these views into a single representation, hence ignoring the unique features of each view and facing challenges with high-dimensional data. The strength of MVL is that it outperforms single-view methods by making use of complementary information from different sources. In MVL, a separate function is learned for each view, and the main goal is to optimize them together for a unified result. According to \cite{tang2020cgd}, MVL models are mainly grouped into three categories: co-training algorithms, co-regularization algorithms, and margin consistency algorithms. Co-training improves robustness by building agreement between views. Co-regularization reduces differences across views during training. Margin consistency ensures stable separations between classes.
Many methods have been developed under these categories. For example, cross-view graph diffusion (CGD) refines graphs for each view through iterative updates and merges them into a unified graph for clustering \cite{tang2020cgd}. SVM-2K combines canonical correlation with SVM to optimize two views jointly \cite{farquhar2005two}. Multi-view twin SVM introduces quadratic programming for non-parallel hyperplanes and adds co-regularization for agreement between views \cite{xie2015multi}. Deep multi-view multiclass twin SVM integrates neural networks and auto-encoders for multiclass tasks, applying similarity-based rules to fuse views \cite{xie2023deep}. MVL has also been applied to different neural architectures, such as a multiview RVFL network that combines multiple views with hidden features for protein prediction \cite{quadir2024multiview}. Predictive subspace learning uses Markov networks to construct latent spaces, assuming conditional independence given hidden variables \cite{chen2010predictive}. 
GRVFL-MV integrates RVFL with MVL and graph embedding to capture both nonlinear relationships and geometric structures for classification \cite{tanveer2025grvfl}. Other applications include multi-view BLS for decoding brain decisions from neural signals \cite{shi2020multi}, and stacking-based multi-view BLS with residual connections, which improves RGB image classification through transfer learning \cite{huang2024stacking}. Several studies provide MVL approaches, covering alignment and fusion methods, from generative models to neural architectures, with applications in classification and retrieval \cite{li2018survey, yu2025review}. The multiview large margin distribution machine models view-specific means and variances to maintain margin distribution while satisfying both consistency and complementarity requirements \cite{hu2024multiview}.The universal multiview dictionary learns shared and class-specific components for action recognition, while using sparsity and locality to preserve feature relationships \cite{yao2017learning}. MVL has become a widely adopted approach for clustering, prediction, and data fusion tasks \cite{yu2025review,haris2024breaking}.

\subsection{Intuitionistic Fuzzy Membership Scheme}
Intuitionistic Fuzzy Sets (IFS) \cite{atanassov1999intuitionistic} handle noise and outliers by using membership, non-membership, and an intuitionistic fuzzy index for each sample. This representation improves decision support in classification and clustering, under the noisy conditions \cite{gao2024interactive,gohain2023distance}. For a sample $(v_i, t_i)$, we compute the membership degree by mapping the data into a high-dimensional kernel feature space. The value is based on the distance between the sample and the corresponding class center, as defined below:
\begin{equation}
\label{eq:7}
\xi(v_i)=
\begin{cases}
1 - \dfrac{\|\kappa(v_i)-C^+\|}{R^+ + \epsilon}, & t_i=+1, \\[8pt]
1 - \dfrac{\|\kappa(v_i)-C^-\|}{R^- + \epsilon}, & t_i=-1,
\end{cases}
\end{equation}
where $C^+$ and $C^-$ denote the centers of the positive and negative classes, $R^+$ and $R^-$ represent their corresponding radii, $\epsilon > 0$ is a small constant to prevent division by zero, and $\kappa(\cdot)$ denotes the kernel mapping function.

The non-membership degree measures how different a sample is from its neighborhood and is defined as:
\begin{equation}
\omega(v_i) = (1 - \xi(v_i))\,\Pi(v_i),
\end{equation}
where $\Pi(v_i)$ represents the proportion of dissimilar data points in the neighborhood of $v_i$.
\begin{equation}
\label{eq:9}
\Pi(v_i) = 
\frac{\;\big|\{v_j:\ \|\kappa(v_i)-\kappa(v_j)\|\le\sigma\}\;\cap\;\{v_j:\ t_j \neq t_i\}\big|\;}
{\;\big|\{v_j:\ \|\kappa(v_i)-\kappa(v_j)\|\le\sigma\}\big|\;}.
\end{equation}
Here $\sigma>0$ is the neighborhood radius and $|\cdot|$ denotes set cardinality. The IF score combines the membership and non-membership values of each sample and is defined as follows:
\begin{equation}
\label{eq:10}
j(v_i) =
\begin{cases}
\xi(v_i), & \omega(x_i)=0, \\[6pt]
0, & \xi(v_i)\le \omega(v_i), \\[6pt]
\dfrac{1-\omega(v_i)}{2 - \xi(v_i) - \omega(v_i)}, & \text{otherwise}.
\end{cases}
\end{equation}
Finally, the diagonal score matrix $J$ is defined as:

\begin{equation}
\label{eq:11}
J = 
\begin{bmatrix}
j(v_1) & \cdots & 0 \\
\vdots & \ddots & \vdots \\
0 & \cdots & j(v_N)
\end{bmatrix}_{N \times N}.
\end{equation}

\subsection{Graph Embedding Strategy}

The Graph Embedding (GE) framework \cite{yan2006graph,nie2020unsupervised} maps graph structure into a vector space, producing representations that can be directly used for computation and analysis while preserving key properties of the graph. Given a training dataset $V$, we construct two graphs: the intrinsic graph $\mathcal{G}^{in} = \{V, W^{in}\}$ and the penalty graph $\mathcal{G}^{p} = \{V, W^{p}\}$.
Here,  $W^{in} \in \mathbb{R}^{N \times N}$ is the similarity weight matrix, 
and $W^{p} \in \mathbb{R}^{N \times N}$ is the penalty weight matrix. 
The optimization problem for GE is formulated as minimizing the trace of 
a projected Laplacian subject to an orthogonality constraint:
\begin{equation}
\begin{aligned}
\hat{c} &= \underset{{\text{tr}(c^T V^T U V c) = f}}{\operatorname{argmin}} \sum_{k \neq l} \left\|c^T v_k - c^T v_l\right\|^2 W_{{kl}}^{in}
\\
  &= \underset{{\text{tr}(c^T V^T U V c) = f}}{\operatorname{argmin}} \text{tr}\left(c^T V^T L V c\right)
 \end{aligned}
\end{equation}
where, $\text{tr}(\cdot)$ denotes the trace operator and $f$ is a constant value. $L = D - W^{in} \in \mathbb{R}^{N \times N}$  is the Laplacian of the intrinsic graph, 
and $U = L_{p} = D_{p} - W^{p}$ is the Laplacian of the penalty graph. 
The diagonal degree matrices $D$ and $D^{p}$ are defined by the row sums 
of $W^{in}$ and $W^{p}$, respectively. The matrix 
$c$ is related to the projection matrix. This ensures that the embedding preserves local structure while pushing that strongly connected nodes closer. The orthogonality constraints ${\text{tr}(c^T V^T U V c) = f}$ stabilize the solution and prevent degeneracy. 
The problem can be reformulated as a generalized eigenvalue system:

\begin{equation}
    H_{in} e = \mu H_{p} e,
\end{equation}

where $H_{in} = V^T L V$, $H_{p} = V^T U V$, $\mu$ denotes the eigenvalue, and $e$ represents the eigenvector basis. The embedding transformation is constructed from the eigenvectors of the matrix $H={H_{p}^{-1}} H_{in}$, here $H$ captures both the intrinsic and penalty graph connections of the data samples.

\section{Proposed Intutionistic Fuzzy Graph embedded Broad Learning System for multi-view learning}
\label{section3}
In this section, we introduce the Multi-View Graph-Embedded Intuitionistic Fuzzy Broad Learning System (MVGIFBLS). Existing BLS methods each address only part of the problem: MVL-based models capture multi-view information but miss structural patterns and struggle with noise, GE-based models preserve geometry but fail under noisy conditions, and IF-based models handle noise but ignore multi-view diversity and structure. Existing BLS-based methods do not simultaneously address these limitations. We propose the MVGIFBLS framework, which incorporates graph embedding, intuitionistic fuzzy theory, and multi-view learning into the BLS architecture to address challenges such as noise, outliers, geometric structure, and relationships among data samples. We incorporate IF sets to reduce the influence of noise and outliers in real-world data by assigning appropriate weights to training samples. GE is integrated into the BLS framework for two main reasons. First, the graph models the neighborhood relationships among data samples, helping preserve the intrinsic geometric structure of the data. Second, the graph embedding transforms this structural information into feature representations that can be effectively utilized by the BLS framework. To further preserve the local structural and neighborhood relationships, Local Fisher Discriminant Analysis (LFDA) \cite{sugiyama2007dimensionality} is incorporated into the graph embedding process, helping maintain the topological properties of the data during learning. Furthermore, the MVL strategy enables the model to jointly learn from multiple feature views, allowing complementary information from different feature representations to be integrated. By combining multi-view learning, graph embedding, and intuitionistic fuzzy theory within the BLS framework, the proposed MVGIFBLS improves the classification performance of BLS on the evaluated benchmark datasets. The optimization formulation of the proposed model is given below:
\begin{align}
\label{eq:14}
\min_{\zeta_1,\zeta_2} 
& \sum_{i=1}^{2} \frac{c_i}{2} ||J_i^{1/2} \theta_i||^2_2 
+ \frac{c_3}{2}||\zeta_1||^2_2 + \frac{c_4}{2}||\zeta_2||^2_2
+ \sum_{i=1}^{2} \frac{\gamma_i}{2} \|H_i^{1/2} \zeta_i\|_2^2 
+ \rho \theta_1^T J_1^{1/2} J_2^{1/2} \theta_2 \nonumber\\
\text{s.t.}  
&\quad Q_1 \zeta_1 - T_{\text{true}} = \theta_1, \quad 
Q_2 \zeta_2 - T_{\text{true}} = \theta_2,
\end{align}
where 
\begin{align}
Q_1 = \begin{bmatrix} Y_E^g & Z_E^h  \end{bmatrix}, \quad
Q_2 = \begin{bmatrix} Y_F^g & Z_F^h \end{bmatrix}, 
\label{eq:15}
\end{align}
$Q_1$ and $Q_2$ are the feature mapping matrices for different views.  The objective function of the proposed MVGIFBLS model includes several terms, each with a specific role. The term 
$\frac{c_i}{2} \|J_i^{1/2} \theta_i\|_2^2$
penalizes the prediction error for the $i^{\text{th}}$ view, where the error is weighted by intuitionistic fuzzy scores to give more importance to reliable samples. The terms 
$\frac{c_3}{2} \| \zeta_1 \|_2^2$ and 
$\frac{c_4}{2} \| \zeta_2 \|_2^2$ 
regularize the output weights, which helps to prevent overfitting and improve generalization. The graph embedding term 
$\frac{\gamma_i}{2} \|H_i^{1/2} \zeta_i\|_2^2$ 
is used to preserve local geometric and structural information in each view, ensuring that similar samples remain close in the learned space. The cross-view consistency term $\rho \, \theta_1^T J_1^{1/2} J_2^{1/2} \theta_2$ encourages both views to produce similar error patterns, promoting agreement between the predictions from different perspectives. $\zeta_1$ and $\zeta_2$ represent the output weight vectors, while $J_1$ and $J_2$ are diagonal matrices containing intuitionistic fuzzy weights. The variables $\theta_1$ and $\theta_2$ denote the error terms between the predicted and actual outputs. Parameters $c_1, c_2, c_3,$ and $c_4$ are regularization coefficients that control the trade-off between minimizing errors and maintaining weight magnitudes. $H_{1}$ and $H_{2}$ refer to graph Laplacian or adjacency matrices that capture the local geometric structure, with $\gamma_1$ and $\gamma_2$ as their respective balancing factors. The parameter $\rho$ ensures consistency among the views. 

The Lagrangian of optimization problem \eqref{eq:14} is: 

\begin{multline}
    \mathcal{L} = \sum_{i=1}^{2} \frac{c_i}{2} ||J_i^{1/2} \theta_i||^2_2 
+ \frac{c_3}{2}||\zeta_1||^2_2 + \frac{c_4}{2}||\zeta_2||^2_2
+ \sum_{i=1}^{2} \frac{\gamma_i}{2} \|H_i^{1/2} \zeta_i\|_2^2 
+ \rho \theta_1^T J_1^{1/2} J_2^{1/2} \theta_2 \\-  \sum_{i=1}^{2}\lambda_i^T (Q_i \zeta_i - T_{\text{true}} - \theta_i).
\end{multline}
The Lagrangian is differentiated with respect to 
$\theta_1$, $\theta_2$, $\zeta_1$, $\zeta_2$, $\lambda_1$, and $\lambda_2$ as shown below:

\begin{align}
\label{eq:17}
\frac{\partial\mathcal{L}}{\partial \theta_1} &= c_1 J_1 \theta_1 + \rho {(J_1 J_2)}^{1/2} \theta_2 + \lambda_1 = 0, \\
\label{eq:18}
\frac{\partial\mathcal{L}}{\partial \theta_2} &= c_2 J_2 \theta_2 + \rho {(J_2 J_1)}^{1/2} \theta_1 + \lambda_2 = 0, \\
\label{eq:19}
\frac{\partial\mathcal{L}}{\partial \zeta_1} &= c_3 \zeta_1 + \gamma_1 H_1 \zeta_1 - Q_1^T \lambda_1 = 0, \\
\label{eq:20}
\frac{\partial\mathcal{L}}{\partial \zeta_2} &= c_4 \zeta_2 + \gamma_2 H_2 \zeta_2 - Q_2^T \lambda_2 = 0, \\
\label{eq:21}
\frac{\partial\mathcal{L}}{\partial \lambda_1} &= Q_1 \zeta_1 - T_{true} - \theta_1 = 0, \\
\label{eq:22}
\frac{\partial\mathcal{L}}{\partial \lambda_2} &= Q_2 \zeta_2 - T_{true} - \theta_2 = 0.
\end{align}
Substituting $(\lambda_1, \lambda_2)$ from \eqref{eq:17}-\eqref{eq:18} into \eqref{eq:19} and \eqref{eq:20}, we get
\noindent
\begin{align}
\label{eq:23}
&c_{3}\zeta_{1} + \gamma_{1}H_{1}\zeta_{1} + Q_{1}^{T}\left(c_{1}J_{1}\theta_{1} + \rho(J_{1}J_{2})^{1/2}\theta_{2}\right) = 0\\
\label{eq:24}
&c_{4}\zeta_{2} + \gamma_{2}H_{2}\zeta_{2} + Q_{2}^{T}\left(c_{2}J_{2}\theta_{2} + \rho(J_{2}J_{1})^{1/2}\theta_{1}\right) = 0
\end{align}
Substituting the value of $(\theta_1, \theta_2)$ from \eqref{eq:21}-\eqref{eq:22} into \eqref{eq:23} and \eqref{eq:24}, we get
\begin{align}
\label{eq:26}
&c_3 \zeta_1 + \gamma_1 H_1 \zeta_1 + Q_1^T 
\Big[c_1 J_1 ( Q_1 \zeta_1 - T_{\text{true}}) + 
\rho (J_1 J_2)^{1/2} ( Q_2 \zeta_2 - T_{\text{true}})
\Big]=0\\
\label{eq:27}
&c_4 \zeta_2 + \gamma_2 H_2 \zeta_2 + Q_2^T 
\Big[
c_2 J_2 ( Q_2 \zeta_2 - T_{\text{true}} ) + 
\rho (J_2 J_1)^{1/2} \left( Q_1 \zeta_1 - T_{\text{true}} \right)
\Big]=0
\end{align}
Simplifying these equations, we get:
\begin{multline}
\Big(c_3 I_1 + \gamma_1 H_1 + c_1 Q_1^T J_1 Q_1 \Big)\zeta_1 +\Big( \rho Q_1^T (J_1 J_2)^{1/2} Q_2\Big) \zeta_2 =  Q_1^T \\\Big(c_1J_1 + \rho(J_1 J_2)^{1/2}\Big) T_{\text{true}}
\end{multline}
\begin{multline}
\Big(\rho Q_2^T (J_2 J_1)^{1/2} Q_1\Big) \zeta_1 + \Big(c_4 I_2 + \gamma_2 H_2 + c_2 Q_2^T J_2 Q_2\Big) \zeta_2   =  Q_2^T \\\Big(c_2 J_2  + \rho (J_2 J_1)^{1/2}\Big) T_{\text{true}} 
\end{multline}
where, \(I_1\) and \(I_2\) are identity matrices with appropriate sizes.   \\
Assuming:
\begin{align*}
A_{11} &= c_3 I + \gamma_1 H_1 + c_1 Q_1^T J_1 Q_1 \\
A_{12} &= \rho Q_1^T (J_1 J_2)^{1/2} Q_2 \\
A_{21} &= \rho Q_2^T (J_2 J_1)^{1/2} Q_1 \\
A_{22} &= c_4 I + \gamma_2 H_2 + c_2 Q_2^T J_2 Q_2
\end{align*}
The system of equations can be expressed in a matrix format as follows:
\begin{equation}
\label{eq:29}
\begin{bmatrix}
\zeta_1 \\
\zeta_2
\end{bmatrix}
=
\begin{bmatrix}
A_{11}
& A_{12} \\
A_{21}
& A_{22}
\end{bmatrix}^{-1}
\begin{bmatrix}
Q_1^T \left(c_1J_1 + \rho(J_1 J_2)^{1/2}\right)\\
Q_2^T \left(c_2 J_2  + \rho   (J_2 J_1)^{1/2}\right) 
\end{bmatrix}
T_{true}
\end{equation}
For a test sample $v$, which has two representations $V_E$ and $V_F$ (from views $E$ and $F$), we define three decision functions:
\begin{align}
\label{eq:30}
& P_{c1} = \text{sign}\!\left(\tfrac{1}{2}\big([v_E \hspace{0.5em} \phi(v_E W_E + b_E)]\zeta_1 + [v_F \hspace{0.5em} \phi(v_F W_F + b_F)]\zeta_2\big)\right), \\
\label{eq:31}
& P_{c2} = \text{sign}\!\big([v_E \hspace{0.5em} \phi(v_E W_E + b_E)]\zeta_1 \big),  \\
\label{eq:32}
& P_{c3} = \text{sign}\!\big([v_F \hspace{0.5em} \phi(v_F W_F + b_F)]\zeta_2 \big).
\end{align}

Here, $P_{c2}$ and $P_{c3}$ represent the prediction results from view~1 and view~2, 
while $P_{c1}$ denotes the combined prediction obtained from both views. 
$W_E$ and $W_F$ are randomly initialized weight matrices, and $b_E$ and $b_F$ 
are the bias vectors taken from the bias matrices $B_E$ and $B_F$, respectively. 
Since all columns of $B_E$ are identical, any column can be chosen as $b_E$, 
and the same applies to $b_F$. The complete procedure of the proposed 
MVGIFBLS method is described in Algorithm \ref{alg:1}.

\begin{algorithm}[H]
\caption{Training steps of the Proposed MVGIFBLS Model}
\label{alg:1}
\textbf{Input:} Training datasets $V_E$ and $V_F$.\\
\textbf{Output:} Optimized output weights
\begin{algorithmic}[1]
\State $c_i, \; n_i, \; \mu, \; \rho, \; \gamma_j \quad \text{where } i = 1,2,\ldots,4 \text{ and } j = 1,2.$

\State Find the values of $Q_1$ and $Q_2$ using equation~\eqref{eq:15}.
\State For Graph Embedding, calculate the following:
\begin{itemize}
    \item[a)] Calculate the intrinsic and penalty graph weights using equations~\eqref{eq:34} and~\eqref{eq:35} for $Q_1$ and $Q_2$.
    
    \item[b)] Find Laplacian matrices: $L_i = D_i - W_i^{\text{in}}$ and $U_i = L_i^{p} = D_i^{p} - W_i^{\text{p}}$ for $Q_i (i=1,2)$.
    
    \item[c)] Calculate: $G_{\text{in}}^1 = Q_1^\top L_1 Q_1$, $G_{\text{p}}^1 = Q_1^\top U_1 Q_1$ and $G_{\text{in}}^2 = Q_2^\top L_2 Q_2$, $G_{\text{p}}^2 = Q_2^\top U_2 Q_2$, for $Q_1$ and $Q_2$, respectively.

    \item[d)]  Find: $H_1= (H_{\text{p}}^1)^{-1} H_{\text{in}}^1$ and $H_2= (H_{\text{p}}^2)^{-1} H_{\text{in}}^2$.
\end{itemize}

\State For Intuitionistic fuzzy, calculate the following:
    \begin{itemize}
        \item[a)] Compute the Membership value $\xi(v_i)$ for $v_i$ using equation~\eqref{eq:7}.
        \item[b)] Compute the Non-membership value $\Pi(v_i)$ for $v_i$ using equation~\eqref{eq:9}.
        \item[c)] Compute the Intuitionistic fuzzy score value $j(v_i)$ using equation~\eqref{eq:10}.
        \item[d)] Calculate the Score matrix $J$ using equation~\eqref{eq:11}.
   \end{itemize}
\State Compute the output weight matrices $\zeta_1$ and $\zeta_2$ using~\eqref{eq:29}.
\State Apply the test conditions in \eqref{eq:30}--\eqref{eq:32} to classify new data points.
\end{algorithmic}
\end{algorithm}

\subsection{Local Fisher Discriminant Learning with Graph Embedding}

The intrinsic and penalty graphs are constructed using the augmented matrices $Q_1$ and $Q_2$. Particularly, for $Q_1$, we have $\mathcal{G}_1^{in}=\{Q_1, W_1^{in}\}$ and $\mathcal{G}_1^{p} = \{Q_1, W_1^{p}\}$. Similarly, for $Q_2$, we have $\mathcal{G}_2^{in}=\{Q_2, W_2^{in}\}$ and $\mathcal{G}_2^{p}=\{Q_2, W_2^{p}\}$. Thus, the intrinsic graphs are $G_{in}^1 = Q_1^\top L_1 Q_1$ and $G_{in}^2 = Q_2^\top L_2 Q_2$, while the penalty graphs are $G_{p}^1 = Q_1^\top U_1 Q_1$ and $G_{p}^2 = Q_2^\top U_2 Q_2$. Within the GE framework, we use the weighting strategy of Local Fisher Discriminant Analysis (LFDA) \cite{sugiyama2007dimensionality} to compute the weights for intrinsic and penalty graphs.
The intrinsic and penalty weights are given as follows:

\begin{align}
_iW_{kl}^{{in}} &= 
\begin{cases}
\dfrac{\eta_{kl}}{n_{t_k}}, & t_k = t_l, \\
0, & \text{otherwise},
\end{cases} \label{eq:34} \\[10pt]
_iW_{kl}^{{p}} &= 
\begin{cases}
\dfrac{\eta_{kl}}{N}\Big(1 - \dfrac{1}{n_{t_k}}\Big), & t_k = t_l, \\[8pt]
\dfrac{1}{N}, & \text{otherwise},
\end{cases} \label{eq:35}
\end{align}
% \begin{equation}
% _iW_{in}(i,j) =
% \begin{cases}
% \dfrac{\eta_{ij}}{n_{c_i}}, & c_i = c_j, \\[8pt]
% 0, & \text{otherwise},
% \end{cases} \\
% W_{p}(i,j) =
% \begin{cases}
% \dfrac{\eta_{ij}}{N}\Big(1 - \dfrac{1}{n_{c_i}}\Big), & c_i = c_j, \\[8pt]
% \dfrac{1}{N}, & \text{otherwise},
% \end{cases}
% \end{equation}
where \( i = 1, 2 \). For this setting, $t_k$ denotes the class label of sample $k$, $n_{t_k}$ denotes the number of samples in the class labeled $t_k$, 
and $N$ is the total number of samples. The similarity between samples $k$ and $l$ is defined as

\begin{equation}
\eta_{ij} = \exp \left( - \frac{\| q_k - q_l \|^2}{2\sigma^2} \right),
\end{equation}
where $q_k$, $q_l$ $\in Q_i (i= 1,2)$, and $\sigma$ is a scaling parameter. 
After constructing the graph matrices, the intrinsic and penalty structures are jointly embedded and optimized along with the final output layer. This joint optimization allows the BLS to exploit both discriminative information and structural relationships, thereby increasing robustness to noise and enhancing the overall representational capacity of the model.

\section{Experiments and Result Discussion}
\label{section4}
In this section, we present the experimental setup, evaluation results, and analysis of the proposed MVGIFBLS model. We describe the datasets used from the UCI~\cite{asuncion2007uci} and 
KEEL~\cite{derrac2015keel} repositories, followed by comparisons with eight baseline models. We further evaluate the robustness of the proposed framework using Gaussian feature noise experiments and conduct an ablation study to analyze the contribution of its key components. The experimental results are supported by comprehensive statistical analyses, including Friedman ranking, Wilcoxon signed-rank test, and win–tie–loss analysis. Finally, we perform a sensitivity analysis of hyperparameters to examine the robustness and stability of the model.  We evaluated the proposed MVGIFBLS model in comparison to eight baseline models: ELM1 \cite{huang2006extreme} (ELM on ‘view – E’), ELM2 \cite{huang2006extreme} (ELM on ‘view – F’), BLS1 \cite{chen2017broad} (BLS on ‘view – E’), BLS2 \cite{chen2017broad} (BLS on ‘view – F’), IFBLS1 \cite{sajid2024intuitionistic} (intuitionistic fuzzy BLS on ‘view – E’), IFBLS2 \cite{sajid2024intuitionistic} (intuitionistic fuzzy BLS on ‘view – F’), MVLDM \cite{hu2024multiview}, and MVRVFL \cite{quadir2024multiview}. A detailed theoretical and empirical comparison with these baselines is presented in the results subsection.

% \subsection{Time Complexity}

% Let us consider a dataset comprising $N$ samples. In the Intuitionistic Fuzzy Broad Learning System (IF-BLS), both membership and non-membership degrees are evaluated for each sample, which introduces a computational cost of $\mathcal{O}(N)$. Nevertheless, the primary computational burden stems from the matrix inversion step. Following the conventional procedure, the inversion of an $N \times N$ matrix requires $\mathcal{O}(N^{3})$ operations. For sufficiently large $N$, the linear term $\mathcal{O}(N)$ becomes negligible in comparison to the cubic complexity, thereby yielding an overall complexity of $\mathcal{O}(N^{3})$ for IF-BLS. 

% In the case of the Fuzzy Broad Learning System (F-BLS), only membership values are computed, which again requires $\mathcal{O}(N)$ operations. Since this term is also dominated by the matrix inversion step, the overall computational complexity of F-BLS remains $\mathcal{O}(N^{3})$.

\subsection{Experimental Setup}
This study proposes an enhanced GE framework that incorporates intuitionistic fuzzy weights into BLS to improve MVL, especially for handling noise and outliers in UCI and KEEL datasets. The similarity between two feature vectors, $z$ and $y$, is measured using a Gaussian kernel defined as:
\[
K(z, y) = \exp\!\left(-\frac{\|z-y\|^{2}}{\mu^{2}}\right),
\]
where $\mu > 0$ is a parameter that directly affects the model performance. All experiments are conducted on an Intel Xeon(R) w5-2565X CPU running at 4.7~GHz with 36 cores, using Ubuntu 22.04.4 LTS and Python~3.9. Benchmark datasets obtained from the UCI \cite{asuncion2007uci} and KEEL \cite{derrac2015keel} repository, covering a diverse range of classification tasks. Each dataset divided into $70\%$ for training and $30\%$ for testing to ensure balanced evaluation.
% Since the total search space is very large (56,357,850 combinations), we used a random search strategy \cite{bergstra2012random} with 50,000 iterations and a fixed seed of 42 for all BLS models,
We adopted a random search strategy with a fixed seed, as \cite{bergstra2012random} demonstrated that random search is more efficient than grid search for hyperparameter optimization in neural network models. This approach allows us to effectively explore the parameter space with fewer iterations while maintaining good performance. We used five-fold cross-validation for hyperparameter optimization. In a five-fold validation procedure, the dataset is divided into five subsets; training is performed on each of the four subsets while the remaining subset is used for validation, and the process is repeated. In our experiments, the regularization parameters $c_i$ $(i = 1,2,3,4)$, the graph regularization parameters $\gamma_j$ $(j = 1,2)$, and the cross-view consistency parameter $\rho$ are tuned within the range $\{10^{-5}, 10^{-4}, \dots, 10^{5}\}$. To simplify the setup, we use a common value for the parameters $c_1, c_2, c_3,$ and $c_4$, and we also set $\gamma_1$ equal to $\gamma_2$. In the BLS models, we set the number of feature nodes per window ($n_1$) to vary from 5 to 50 in steps of 5, the number of feature groups ($n_2$) to take values from 1 to 21 with a step of 2, the number of enhancement nodes ($n_3$) to range from 5 to 105 in steps of 10, and the number of enhancement groups ($n_4$) fixed at 1. For ELM and MVRVFL, the regularization parameter is selected from the same range, while the number of hidden neurons $(n)$ varies from $5$ to $105$ with a step size of $10$. In addition, the kernel bandwidth parameter $\mu$ is selected from the values $\{0.01, 0.1, 0.5, 1.0, 2.0, 5.0, 10.0\}$.

\subsection{Results on UCI and KEEL Datasets}
In this study, we used datasets from two sources: the UCI \cite{asuncion2007uci} and the KEEL \cite{derrac2015keel} machine learning dataset repository. The UCI repository provides datasets from areas such as health, science, and finance, and is often used for comparing machine learning methods. The KEEL repository, on the other hand, is mainly an imbalanced data repository. It includes both original and preprocessed datasets for classification and regression problems. Since the UCI and KEEL datasets are inherently single-view datasets, we construct a synthetic second view for each dataset. The original feature representation is treated as view-E, while the synthetic view (view-F) is generated by applying Principal Component Analysis (PCA) and retaining the top 95\% principal components, following the experimental protocol adopted in \cite{arora2026robust}. We compare the performance of the proposed MVGIFBLS model with eight baseline models  (ELM1, ELM2, BLS1, BLS2, IFBLS1, IFBLS2, MVLDM, and MVRVFL). Table \ref{table:results1} shows the AUC values of different models on various datasets, along with the average AUC and the ranking of each model based on the UCI and KEEL datasets.

The performance of the proposed MVGIFBLS model is evaluated on 13 benchmark datasets and compared with eight baseline models given in Table \ref{table:results1}. The comparison is carried out using the Area Under the Curve (AUC) metric, which provides a reliable measure of classification performance, especially for imbalanced datasets. From the results, it is clear that the proposed MVGIFBLS model consistently performs well across most datasets. It obtains the lowest average rank of 2.88 among all methods. In comparison, MVRVFL and ELM1 have average ranks of 4.00 and 4.12, respectively, while the other methods show higher ranks. Across individual datasets, MVGIFBLS records the highest AUC values on several cases. For instance, it achieves top results on breast-cancer-wisc-diag (98.44\%), cylinder-bands (70.18\%), horse-colic (68.58\%), musk-1 (64.7\%), and spect (71.88\%). It also performs well on datasets such as pima (75.70\%) and statlog-german-credit (71.9\%), where its results are close to the top methods.

\begin{table*}[h]
% \vspace*{-60pt}
\caption{Performance comparison of baseline models and the proposed MVGIFBLS model on UCI and KEEL datasets}
\centering
\label{table:results1}
\begin{adjustwidth}{-3.0cm}{4.5cm}
\setlength{\tabcolsep}{4pt} 
\renewcommand{\arraystretch}{1.6}
\resizebox{1.6\linewidth}{!}{%
\begin{tabular}{|c|c|c|c|c|c|c|c|c|c|}
\hline
 & \textbf{ELM1~\cite{huang2006extreme}} & \textbf{ELM2~\cite{huang2006extreme}} & \textbf{BLS1~\cite{chen2017broad}} & \textbf{BLS2~\cite{chen2017broad}} & \textbf{IFBLS1~\cite{sajid2024intuitionistic}} & \textbf{IFBLS2~\cite{sajid2024intuitionistic}} &
 \textbf{MVLDM~\cite{hu2024multiview}} &
 \textbf{MVRVFL~\cite{quadir2024multiview}} &
\textbf{MVGIFBLS}$^*$\\\hline
\textbf{Dataset}  & \textbf{AUC~$\uparrow$} & \textbf{AUC~$\uparrow$} & \textbf{AUC~$\uparrow$} &
\textbf{AUC~$\uparrow$} & \textbf{AUC~$\uparrow$} & \textbf{AUC~$\uparrow$} &
\textbf{AUC~$\uparrow$}&
\textbf{AUC~$\uparrow$}&
\textbf{AUC~$\uparrow$} \\
$(\text{Patterns} \times \text{Features})$&
$(n,c)$ &
$(n,c)$ &
$(n_1,n_2,n_3,n_4,c)$ &
$(n_1,n_2,n_3,n_4,c)$ &
$(n_1,n_2,n_3,n_4,c,\mu)$ &
$(n_1,n_2,n_3,n_4,c,\mu)$ &
$(d, \epsilon,\lambda, v1, v2, \theta)$&
$(n, c1, c3, \rho)$&
$(n_1,n_2,n_3,n_4,c,\mu,\gamma,\rho)$\\
\hline

breast-cancer-wisc-diag & 95.47 & 96.40 & 93.75 & 81.25 & 89.84 & 91.41 & 96.88 & 91.23 & \textbf{98.44} \\
$(569\times30)$ & $(105,1)$ & $(105,10^{-1})$ & $(40,21,85,1,1)$ & $(45,3,25,1,1)$ & $(10,17,5,1,10,32)$ & $(45,5,75,1,10,8)$ & $(1,10^{-2},2,-2,-2,1)$ & $(95,10^{2},10^{-5},10^{2})$ & $(30,19,95,1,10^{2},10^{-3},1,2)$ \\\hline

cmc & 68.31 & 68.16 & 66.21 & 71.46 & 70.93 & 64.76 & 69.72 & \textbf{71.95} & 70.47 \\
$(1473\times9)$ & $(35,10^{-1})$ & $(55,10^{4})$ & $(40,21,55,1,1)$ & $(20,11,35,1,1)$ & $(35,5,75,1,1,32)$ & $(30,5,15,1,1,32)$ & $(10,10^{-2},2,-2,2,-1)$ & $(105,1,10^{-5},1)$ & $(35,9,85,1,10^{3},10^{3},10^{-5},0.5)$ \\\hline

cylinder-bands & 64.80 & 58.74 & 65.35 & 67.78 & 56.97 & 53.71 & 48.65 & 59.74 & \textbf{70.18} \\
$(512\times35)$ & $(105,10^{-1})$ & $(85,10^{4})$ & $(15,19,45,1,1)$ & $(35,5,85,1,10^{2})$ & $(20,11,5,1,10^{-2},8)$ & $(30,19,55,1,10,32)$ & $(1,10^{-2},-2,-2,-2,-1)$ & $(105,10^{3},10,10^{3})$ & $(5,17,105,1,10^{-3},10^{-2},10^{-5},16)$ \\\hline

horse-colic & 60.13 & 66.83 & 64.43 & 67.46 & 64.91 & 62.28 & 66.43 & 51.11 & \textbf{68.58} \\
$(300\times25)$ & $(55,10^{-5})$ & $(65,10^{-4})$ & $(35,3,85,1,10^{4})$ & $(35,7,75,1,10^{3})$ & $(15,17,25,1,10,4)$ & $(40,3,95,1,10,4)$ & $(50,10^{-2},-2,-2,2,2)$ & $(85,10^{2},10^{-5},10^{2})$ & $(40,3,25,1,10,10,10^{-4},1)$ \\\hline

molec-biol-promoter & 75 & 71.88 & \textbf{81.25} & 71.88 & \textbf{81.25} & 62.5 & \textbf{81.25} & 71.88 & 75 \\
$(106\times57)$ & $(55,10^{-2})$ & $(55,10^{-2})$ & $(40,3,55,1,10^{3})$ & $(20,9,65,1,10^{2})$ & $(40,3,55,1,10^{5},1)$ & $(20,13,25,1,10^{5},0.03125)$ & $(1,10^{-2},-2,1,-2,2)$ & $(15,10^{-5},10^{-5},10^{-5})$ & $(10,21,95,1,10^{2},10,1,0.25)$ \\\hline

musk-1 & 52.45 & 55.11 & 49.76 & 52.18 & 48.77 & 48.95 & 60.33 & 51.75 & \textbf{64.7} \\
$(476\times166)$ & $(45,10^{-3})$ & $(85,10^{-2})$ & $(15,11,45,1,10^{2})$ & $(35,15,65,1,10^{4})$ & $(15,15,65,1,1,0.125)$ & $(20,17,55,1,10,0.125)$ & $(1,10^{-2},-2,-2,-1,2)$ & $(105,10^{-2},10^{-5},10^{-2})$ & $(20,19,45,1,10^{2},10^{-4},10^{-2},0.5)$ \\\hline

oocytes-merluccius-nucleus-4d & 53.44 & 52.36 & 52.79 & 51.26 & 51.02 & 51.53 & 49.72 & \textbf{64.82} & 52.74 \\
$(1022\times41)$ & $(35,1)$ & $(35,10^{-2})$ & $(5,9,15,1,10^{3})$ & $(30,9,35,1,10^{5})$ & $(5,1,25,1,10^{5},16)$ & $(5,9,55,1,10^{-1},4)$ & $(1,10^{-2},-2,-2,2,-1)$ & $(105,10^{3},10^{-5},10^{3})$ & $(35,7,5,1,10^{-1},10^{-2},10^{3},0.25)$ \\\hline

oocytes-trisopterus-nucleus-2f & 50.93 & 51.02 & 49.68& 49.65 & 50 & 49.74 & 37.61 & \textbf{54.74} & 50.43 \\
$(912\times25)$ & $(5,10^{-3})$ & $(5,10^{-1})$ & $(50,11,65,1,10^{-4})$ & $(5,1,5,1,10^{-5})$ & $(50,7,65,1,10^{3},16)$ & $(5,9,45,1,10^{-3},4)$ & $(50,10^{-2},-2,-2,1,-2)$ & $(105,10^{4},10^{3},10^{4})$ & $(40,17,5,1,10^{3},1,10^{5},4)$ \\\hline

pima & \textbf{77.51} & 68.26 & 66.42 & 74 & 70.06 & 75.80 & 69.64 & 69.7 & 75.70 \\
$(768\times8)$ & $(45,10^{2})$ & $(85,10^{-1})$ & $(45,11,5,1,1)$ & $(30,21,15,1,1)$ & $(10,19,45,1,1,8)$ & $(20,17,65,1,1,16)$ & $(1,10^{-2},2,-2,-2,1)$ & $(95,10^{4},10^{3},10^{4})$ & $(35,17,35,1,10^{3},10^{4},10^{-5},0.25)$ \\\hline

pittsburg-bridges-T-OR-D & 42.59 & 50 & 48.15 & 50 & 55.09 & 50 & 48.15 & \textbf{67.74} & 50 \\
$(102\times7)$ & $(85,10)$ & $(55,10^{-2})$ & $(5,1,85,1,10^{-3})$ & $(5,1,35,1,10^{-3})$ & $(45,15,85,1,10^{2},16)$ & $(30,17,5,1,10^{2},8)$ & $(50,10^{-2},2,2,-2,2)$ & $(75,10^{3},10^{-5},10^{3})$ & $(15,21,105,1,10,10^{-5},10^{4},0.25)$ \\\hline

ripley & 90.67 & 91.74 & 89.08 & 87.8 & 86.43 & 89.35 & \textbf{92.00} & 91.2 & 89.35 \\
$(1250\times2)$ & $(105,10^{5})$ & $(35,10^{5})$ & $(30,19,75,1,10^{-1})$ & $(30,15,35,1,10^{-2})$ & $(50,13,25,1,10^{3},0.25)$ & $(20,7,15,1,10^{2},16)$ & $(10^{2},10^{-2},-2,2,2,0)$ & $(105,10^{5},10^{-2},10^{5})$ & $(45,13,105,1,10,10^{2},10^{-2},2)$ \\\hline

spect & 62.5 & 53.13 & 56.25 & 56.25 & 59.38 & 56.25 & 53.13 & 58.33 & \textbf{71.88} \\
$(79\times22)$ & $(55,10^{-5})$ & $(85,10^{-4})$ & $(20,1,15,1,10^{5})$ & $(25,7,5,1,10^{4})$ & $(15,9,55,1,10^{-1},8)$ & $(30,13,45,1,10^{-2},16)$ & $(1,10^{-2},1,-2,-2,2)$ & $(35,10^{-5},10^{-5},10^{-5})$ & $(50,1,85,1,10^{2},10,10,0.0625)$ \\\hline

statlog-german-credit & 72.62 & 65.63 & 66.83 & 65.87 & 60.32 & 66.03 & 63.57 & \textbf{74.67} & 71.90 \\
$(1000\times24)$ & $(105,10^{-3})$ & $(75,10^{-3})$ & $(40,9,5,1,10^{3})$ & $(15,5,95,1,10^{2})$ & $(50,21,105,1,10^{4},0.03125)$ & $(35,13,65,1,10^{5},0.03125)$ & $(10,10^{-2},2,2,2,1)$ & $(105,1,10^{-5},1)$ & $(15,11,45,1,10^{2},10^{2},1,0.25)$ \\\hline

\textbf{Average AUC~$\uparrow$} & 66.65 & 65.33 & 65.38 & 65.10 & 65.00 & 63.25 & 64.39 & 67.60 & \textbf{69.95} \\ \hline
\textbf{Average Rank~$\downarrow$} & 4.12 & 5.08 & 5.73 & 5.42 & 5.77 & 6.31 & 5.69 & 4 & \textbf{2.88} \\ \hline
\multicolumn{6}{l}{Here, $^*$ denotes the proposed model.}
\end{tabular}
}
\end{adjustwidth}
\end{table*}

Although MVGIFBLS does not achieve the highest AUC on every dataset, it maintains consistent performance across all datasets. In contrast, several baseline methods perform well on some datasets but show less consistent performance across the benchmark datasets. ELM and BLS models show more variation in their AUC values, which affects their overall ranking. Overall, the results demonstrate that MVGIFBLS is a reliable and effective multi-view learning model. Its ability to achieve the highest average rank and maintain consistent performance across diverse datasets demonstrates its effectiveness in comparison with existing randomized learning approaches.

\subsection{Results on Animal with Attributes datasets}
\begin{figure}[H]
\centering
\includegraphics[width=0.6\linewidth]{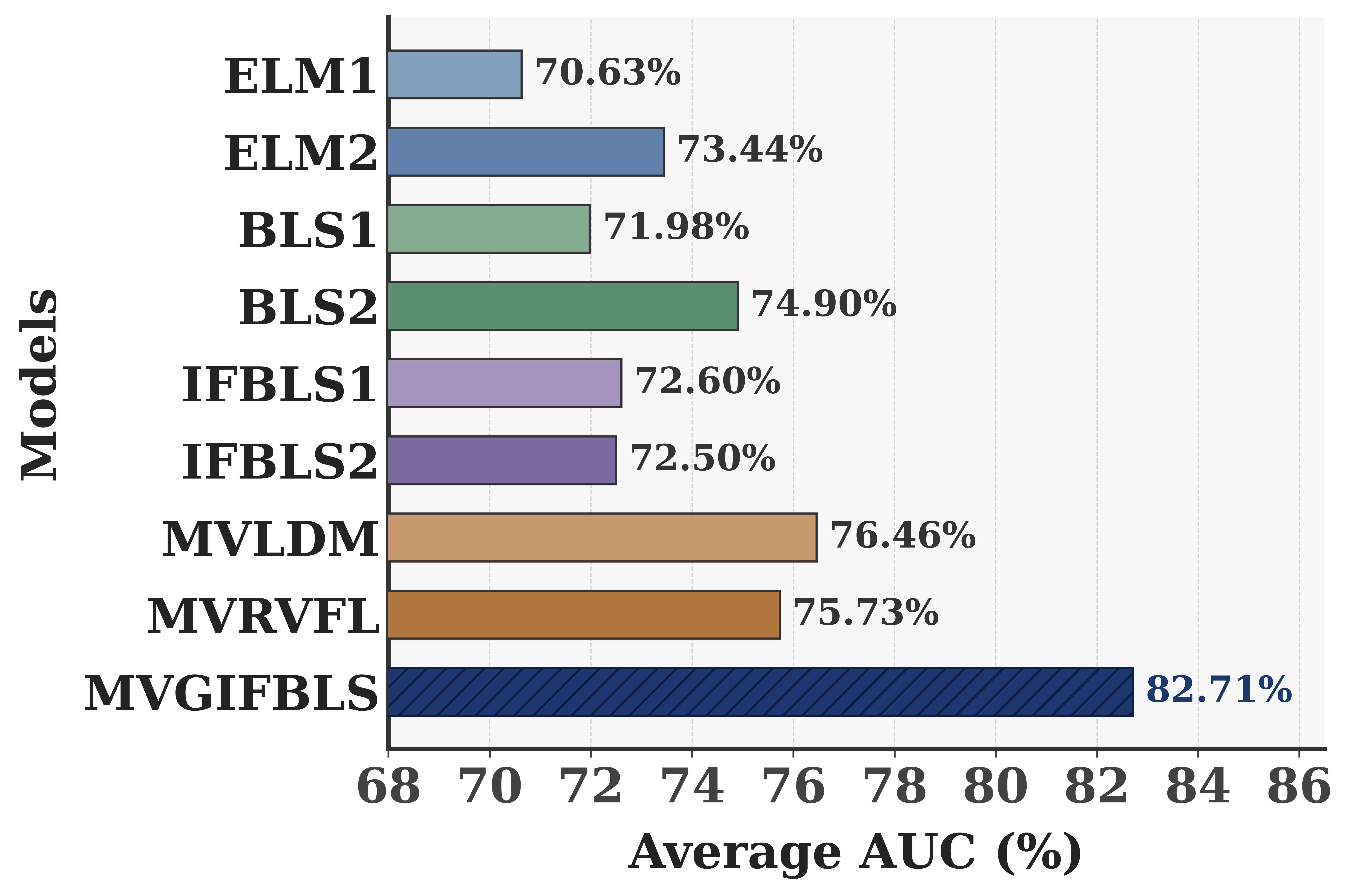}
\caption{Comparison of average AUC values of different models on the AwA dataset.}
\label{fig:auc}
\end{figure}

\begin{figure}[H]
\centering
\includegraphics[width=0.6\linewidth]{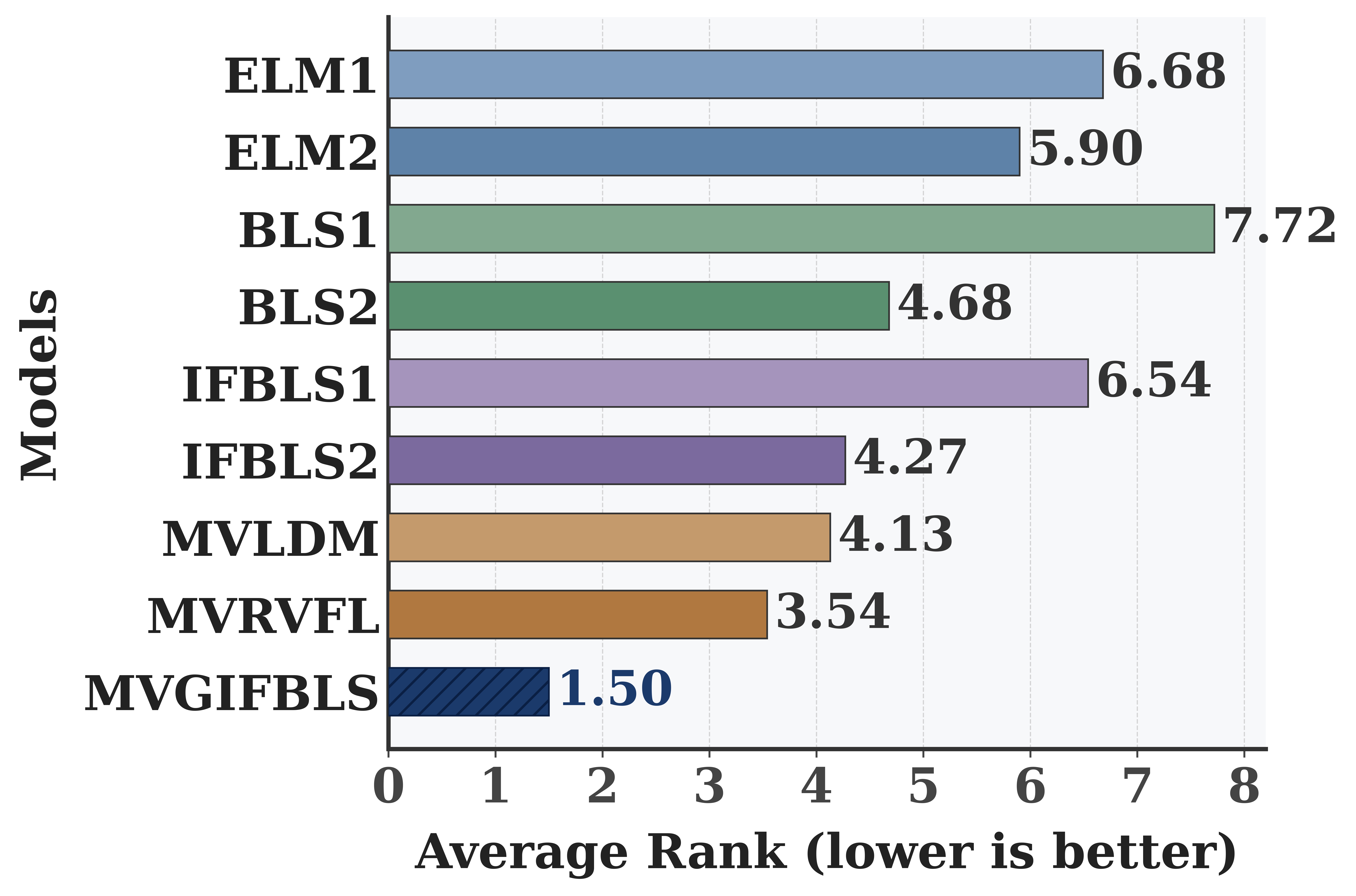}
\caption{Comparison of average rank values of different models on the AwA dataset.}
\label{fig:rank}
\end{figure}

The Animals with Attributes (AwA5) \cite{lampert2013attribute} dataset contains 30,475 images from 50 animal classes, where each image is described using six types of pre-extracted features. In this study, we selected the chimpanzee, giant panda, leopard, Persian cat, pig, hippopotamus, humpback whale, and raccoon classes from the AwA dataset.
Two feature representations are used for analysis: a 252-dimensional Histogram of Oriented Gradients (HOG) descriptor (view-A) and a 2000-dimensional $L_1$-normalized Speeded-Up Robust Features (SURF) descriptor (view-B). Following \cite{tanveer2025grvfl}, binary classification was performed using pairwise combinations of the selected classes.
Table~\ref{tab:awa_results} reports the AUC performance and average rank of different models on the AwA datasets, while Fig.~\ref{fig:auc} and Fig.~\ref{fig:rank} provide a visual comparison. The results show that the proposed MVGIFBLS model achieves the highest overall performance.  The proposed MVGIFBLS model obtains the highest average AUC value of 82.71\%, which is higher than all other compared methods. Among the baseline methods, MVLDM achieves the second-highest average AUC of 76.46\%, followed by MVRVFL with 75.73\%. In comparison, the ELM, BLS, and IFBLS variants achieve lower average AUC values. These results indicate that the proposed method improves the classification performance on multi-view animal datasets. The proposed MVGIFBLS model also achieves the average rank of 1.5, which confirms its stable performance across different class pairs. MVRVFL obtains the second-highest average rank of 3.54, while MVLDM achieves an average rank of 4.13. The remaining methods show higher average ranks, indicating less consistent performance. For individual datasets, the proposed method achieves the highest AUC value of 96.67\% on the chimpanzee\_vs\_Humpbackwhale dataset. It also achieves strong results on chimpanzee\_vs\_Hippopotamus with an AUC of 85\%, and on Persiancat\_vs\_Humpbackwhale and \path{Chimpanzee_vs_Humpbackwhale} with an AUC of 90\%. Similarly, the proposed method achieves better classification performance than most competing methods on datasets such as Pig\_vs\_Hippopotamus, Pig\_vs\_Raccoon, chimpanzee\_vs\_Leopard, and Giantpanda\_vs\_Pig. 

Although some competing methods achieve slightly better performance on a few datasets, their overall performance is less stable. For example, MVRVFL achieves 95\% AUC on the Giantpanda\_vs\_Humpbackwhale dataset, while IFBLS2 obtains 90\% AUC on the chimpanzee\_vs\_Pig dataset. However, these methods do not maintain similar performance across all datasets.

\begin{table*}[h]
\centering
% \vspace*{-70pt}
\caption{Performance comparison of baseline models and the proposed MVGIFBLS model on AwA datasets}
\label{tab:awa_results}
\begin{adjustwidth}{-3.0cm}{4.5cm}
\setlength{\tabcolsep}{4pt} 
\renewcommand{\arraystretch}{1.6}
\resizebox{1.6\linewidth}{!}{%
\begin{tabular}{|c|c|c|c|c|c|c|c|c|c|}
\hline
 & \textbf{ELM1~\cite{huang2006extreme}} & \textbf{ELM2~\cite{huang2006extreme}} & \textbf{BLS1~\cite{chen2017broad}} & \textbf{BLS2~\cite{chen2017broad}} & \textbf{IFBLS1~\cite{sajid2024intuitionistic}} & \textbf{IFBLS2~\cite{sajid2024intuitionistic}} &
 \textbf{MVLDM~\cite{hu2024multiview}} &
 \textbf{MVRVFL~\cite{quadir2024multiview}} &
\textbf{MVGIFBLS}$^*$\\\hline
\textbf{Dataset}  & \textbf{AUC~$\uparrow$} & \textbf{AUC~$\uparrow$} & \textbf{AUC~$\uparrow$} &
\textbf{AUC~$\uparrow$} & \textbf{AUC~$\uparrow$} & \textbf{AUC~$\uparrow$} &
\textbf{AUC~$\uparrow$}&
\textbf{AUC~$\uparrow$}&
\textbf{AUC~$\uparrow$} \\
$(\text{Patterns} \times \text{Features})$&
$(n,c)$ &
$(n,c)$ &
$(n_1,n_2,n_3,n_4,c)$ &
$(n_1,n_2,n_3,n_4,c)$ &
$(n_1,n_2,n_3,n_4,c,\mu)$ &
$(n_1,n_2,n_3,n_4,c,\mu)$ &
$(d, \epsilon,\lambda, v1, v2, \theta)$&
$(n, c1, c3, \rho)$&
$(n_1,n_2,n_3,n_4,c,\mu,\gamma,\rho)$\\
\hline

Chimpanzee\_vs\_Giantpanda&65&66.67&56.67&70&58.33&\textbf{81.67}&73.33&68.33&76.67\\
$\big(200\times(2000\ \&\ 252)\big)$&$(105,10^{-3})$&$(55,10^{-5})$&$(15,21,35,1,10^2)$&$(20,15,45,1,10^2)$&$(15,21,5,1,10,0.25)$&$(20,9,35,1,10^{-4},8)$&$(50,10^{-2},2,-2,1,2)$&$(35,10^{-3},10^{-4},1)$&$(35,21,55,1,10^3,10^{-3},10^2,0.5)$\\\hline

Chimpanzee\_vs\_Hippopotamus&76.67&75&73.33&80&73.33&76.67&81.67&78.33&\textbf{85}\\
$\big(200\times(2000\ \&\ 252)\big)$&$(55,10^{-5})$&$(105,10^{-5})$&$(45,3,65,1,10^4)$&$(25,7,105,1,10^4)$&$(45,3,65,1,10^{-1},3.125\times10^{-2})$&$(20,3,25,1,10^{-1},2)$&$(1,10^{-2},-2,-2,1,2)$&$(5,10^2,1,10^3)$&$(5,19,5,1,1,10^{-4},10^{-5},1)$\\\hline

Chimpanzee\_vs\_Humpbackwhale&90&85&83.33&91.67&86.67&90&91.67&93.33&\textbf{96.67}\\
$\big(200\times(2000\ \&\ 252)\big)$&$(105,10^{-3})$&$(65,10^{-5})$&$(40,3,75,1,10^2)$&$(40,7,5,1,10^2)$&$(40,5,75,1,10^{-5},6.25\times10^{-2})$&$(30,17,35,1,10^5,3.125\times10^{-2})$&$(1,10^{-2},-2,-2,-2,2)$&$(35,10^{-2},1,10^{-1})$&$(50,15,55,1,10^3,10^{-1},10^{-3},10^{-1})$\\\hline

Chimpanzee\_vs\_Leopard&66.67&70&63.33&70&65&73.33&63.33&71.67&\textbf{80}\\
$\big(200\times(2000\ \&\ 252)\big)$&$(95,10^{-5})$&$(115,10^{-3})$&$(50,7,45,1,10)$&$(40,3,25,1,10)$&$(50,7,15,1,10,16)$&$(40,1,25,1,10^{-1},3.125\times10^{-2})$&$(10^2,10^{-2},1,2,1,2)$&$(75,10^{-4},10,1)$&$(20,9,85,1,10^5,10^{-1},10^4,1)$\\\hline

Chimpanzee\_vs\_Pig&75&66.67&60&70&65&\textbf{90}&70&65&80\\
$\big(200\times(2000\ \&\ 252)\big)$&$(95,10^{-5})$&$(95,10^{-4})$&$(15,11,55,1,10^2)$&$(30,11,35,1,10^5)$&$(15,9,75,1,10,6.25\times10^{-2})$&$(25,9,95,1,1,8)$&$(1,10^{-2},-2,-2,2,2)$&$(95,10^{-2},10^2,10^3)$&$(45,21,25,1,10^2,10^{-1},10^{-3},10^{-1})$\\\hline

Chimpanzee\_vs\_Raccoon&61.67&76.67&71.67&73.33&76.67&\textbf{83.33}&73.33&68.33&81.67\\
$\big(200\times(2000\ \&\ 252)\big)$&$(75,10^{-5})$&$(35,10^{-5})$&$(50,17,95,1,10^4)$&$(45,9,85,1,10^5)$&$(50,21,75,1,10^{-3},6.25\times10^{-2})$&$(10,19,35,1,10^{-1},1)$&$(1,10^{-2},-2,2,2,2)$&$(95,10^{-1},10^3,10^2)$&$(35,17,15,1,10^2,10^2,1,2)$\\\hline

Giantpanda\_vs\_Humpbackwhale&91.67&90&88.33&91.67&90&90&93.33&\textbf{95}&93.33\\
$\big(200\times(2000\ \&\ 252)\big)$&$(65,10^{-5})$&$(75,10^{-3})$&$(20,9,65,1,10^2)$&$(40,21,45,1,10^4)$&$(20,13,105,1,10^{-1},16)$&$(30,13,45,1,10^{-2},16)$&$(1,10^{-2},-2,-2,-2,2)$&$(65,10^2,10^3,10^2)$&$(50,15,35,1,10^3,10,10^2,10)$\\\hline

Giantpanda\_vs\_Pig&51.67&60&63.33&66.67&63.33&56.67&68.33&68.33&\textbf{76.67}\\
$\big(200\times(2000\ \&\ 252)\big)$&$(115,10^{-1})$&$(85,10^{-1})$&$(30,13,75,1,10^3)$&$(15,21,35,1,10^5)$&$(25,11,25,1,1,32)$&$(45,13,85,1,10^5,0.5)$&$(1,10^{-2},2,2,-1,2)$&$(55,10^2,10^2,10)$&$(50,21,45,1,10^4,10^3,1,10^{-2})$\\\hline

Leopard\_vs\_Hippopotamus&71.67&76.67&70&71.67&73.33&61.67&71.67&\textbf{81.6}7&80\\
$\big(200\times(2000\ \&\ 252)\big)$&$(55,10^{-2})$&$(45,10^{-3})$&$(10,7,15,1,10)$&$(50,1,5,1,10^2)$&$(45,5,55,1,10^2,8)$&$(10,3,25,1,10^2,1)$&$(50,10^{-2},-2,2,0,2)$&$(75,10^{-1},10^2,10^4)$&$(50,9,45,1,10^5,10^{-5},10^2,1)$\\\hline

Leopard\_vs\_Persiancat&70&73.33&73.33&71.67&71.67&73.33&76.67&80&\textbf{83.33}\\
$\big(200\times(2000\ \&\ 252)\big)$&$(75,10^{-3})$&$(15,10^{-5})$&$(5,21,25,1,10)$&$(15,13,55,1,10^2)$&$(35,11,95,1,10^{-5},8)$&$(20,5,5,1,10^4,0.25)$&$(10^2,10^{-2},-2,2,1,2)$&$(75,10^3,10^4,10)$&$(30,21,75,1,10,10,10^{-3},0.5)$\\\hline

Leopard\_vs\_Pig&51.67&60&63.33&66.67&63.33&71.67&66.67&68.33&\textbf{73.33}\\
$\big(200\times(2000\ \&\ 252)\big)$&$(115,10^{-1})$&$(85,10^{-1})$&$(30,13,75,1,10^3)$&$(15,21,35,1,10^5)$&$(25,11,25,1,1,32)$&$(35,21,75,1,10^{-5},0.125)$&$(1,10^{-2},2,2,-1,2)$&$(75,10^{-2},1,10^4)$&$(30,21,75,1,10^5,10^{-5},10^2,2)$\\\hline

Persiancat\_vs\_Hippopotamus&61.67&78.33&75&73.33&68.33&66.67&\textbf{86.67}&71.67&80\\
$\big(200\times(2000\ \&\ 252)\big)$&$(75,10^{-5})$&$(75,10^{-3})$&$(15,9,85,1,10^2)$&$(15,15,65,1,10^2)$&$(25,7,15,1,10^{-3},16)$&$(35,1,55,1,10^{-4},32)$&$(1,10^{-2},-2,-2,0,2)$&$(5,10^3,10^3,10^{-4})$&$(30,9,55,1,10^3,1,10^2,5)$\\\hline

Persiancat\_vs\_Humpbackwhale&81.67&80&83.33&80&85&63.33&83.33&80&\textbf{90}\\
$\big(200\times(2000\ \&\ 252)\big)$&$(115,10^{-5})$&$(115,10^{-3})$&$(40,15,35,1,10^4)$&$(10,19,105,1,10^4)$&$(40,15,85,1,1,0.125)$&$(10,21,65,1,10,2)$&$(10^2,10^{-2},1,2,2,2)$&$(85,10^{-3},10^{-3},10^{-1})$&$(50,9,5,1,10^{-1},10^{-1},10^{-4},10^{-1})$\\\hline

Pig\_vs\_Hippopotamus&63.33&63.33&63.33&63.33&63.33&60&60&61.67&\textbf{75}\\
$\big(200\times(2000\ \&\ 252)\big)$&$(105,10^{-3})$&$(55,10^{-3})$&$(20,13,75,1,10^3)$&$(50,5,45,1,10^5)$&$(20,13,75,1,10^5,1)$&$(25,7,15,1,10^{-4},32)$&$(1,10^{-2},-2,-2,-1,2)$&$(75,10^{-4},10,1)$&$(50,3,25,1,10^5,10^{-4},10^2,1)$\\\hline

Pig\_vs\_Raccon&70&70&75&75&70&60&70&75&\textbf{81.67}\\
$\big(200\times(2000\ \&\ 252)\big)$&$(115,10^{-2})$&$(95,10^{-3})$&$(15,11,55,1,10)$&$(50,9,45,1,10^3)$&$(20,19,105,1,10^4,0.125)$&$(20,3,5,1,10^4,3.125\times10^{-2})$&$(10,10^{-2},2,-2,1,2)$&$(95,10^{-1},10^3,10^2)$&$(30,13,95,1,10^3,10^{-5},10^2,2)$\\\hline

Raccoon\_vs\_Humpbackwhale&81.67&83.33&88.33&83.33&88.33&61.67&\textbf{93.33}&85&90\\
$\big(200\times(2000\ \&\ 252)\big)$&$(95,10^{-2})$&$(55,10^{-2})$&$(20,9,65,1,10^5)$&$(25,13,75,1,10)$&$(20,9,65,1,10^{-5},0.125)$&$(15,21,65,1,10^{-2},0.5)$&$(1,10^{-2},-2,-2,-2,2)$&$(75,10^{-2},10^{-2},10^{-2})$&$(50,15,45,1,10^5,10^5,1,1)$\\\hline

\textbf{Average AUC~$\uparrow$} & 70.63 & 73.44 & 71.98 & 74.9 & 72.6 & 72.5 & 76.46 & 75.73 & \textbf{82.71} \\ \hline
\textbf{Average Rank~$\downarrow$} & 6.68 & 5.90 & 7.72 & 4.68 & 6.54 & 4.27 & 4.13 & 3.54 & \textbf{1.5} \\ \hline
\hline
\multicolumn{9}{l}{Here, $^*$ denotes the proposed model.}
\end{tabular}
}
\end{adjustwidth}
\end{table*}

In contrast, the proposed MVGIFBLS model consistently achieves high AUC values across most class pairs. Overall, the results demonstrate that the proposed MVGIFBLS model maintains consistent classification performance across different AwA datasets. The combination of multi-view learning, graph embedding, and intuitionistic fuzzy weighting enables the model to integrate information from different feature representations, resulting in improved classification performance.

\subsection{Results on AwA Datasets under Gaussian Feature Noise}

Table~\ref{tab:awa_noise} presents the AUC results of the baseline methods and the proposed MVGIFBLS on the AwA dataset under four levels of Gaussian feature noise (5\%, 10\%, 15\%, and 20\%). The experiments were conducted to evaluate the robustness of the models when the input features are affected by noise. Overall, the proposed MVGIFBLS achieves the highest average AUC of 81.90\%, followed by MVRVFL with 78.33\% and MVLDM with 77.19\%. In comparison, the single-view methods, including ELM1, BLS1, and IFBLS1, obtain average AUC values of 70.86\%, 72.97\%, and 71.88\%, respectively. These results indicate that the proposed multi-view framework provides more consistent classification performance than the compared methods under different noise levels.
The average ranking results also support this observation. MVGIFBLS achieves the average rank of 1.91, while MVRVFL and MVLDM obtain average ranks of 3.66 and 3.88, respectively. The remaining baseline methods have noticeably higher average ranks, indicating relatively lower overall performance.
The proposed MVGIFBLS maintains stable performance under different levels of Gaussian feature noise. The average AUC is 82.19\% at 5\% noise, 81.77\% at 10\% noise, 80.62\% at 15\% noise, and 83.02\% at 20\% noise. These results indicate that the proposed framework remains effective under different levels of Gaussian feature noise.
Overall, the experimental results show that the proposed MVGIFBLS consistently achieves higher AUC values than the compared baseline methods under different levels of Gaussian feature noise on the AwA datasets.

\begin{table*}[!htbp]
\centering
% \vspace*{-90pt}
\caption{Performance comparison of baseline models and the proposed MVGIFBLS model on AwA datasets under different levels of Gaussian feature noise (5\%, 10\%, 15\%, and 20\%).}
\begin{adjustwidth}{-2.8cm}{4.2cm}
\label{tab:awa_noise}
\setlength{\tabcolsep}{4pt} 
\renewcommand{\arraystretch}{1.3}
\resizebox{1.6\linewidth}{!}{%
\begin{tabular}{|c|c|c|c|c|c|c|c|c|c|c|c|}
\hline
 & & \textbf{ELM1~\cite{huang2006extreme}} & \textbf{ELM2~\cite{huang2006extreme}} & \textbf{BLS1~\cite{chen2017broad}} & \textbf{BLS2~\cite{chen2017broad}} & \textbf{IFBLS1~\cite{sajid2024intuitionistic}} & \textbf{IFBLS2~\cite{sajid2024intuitionistic}} &
 \textbf{MVLDM~\cite{hu2024multiview}} &
 \textbf{MVRVFL~\cite{quadir2024multiview}} &
\textbf{MVGIFBLS}$^*$\\\hline
\textbf{Dataset} & \textbf{Noise} & \textbf{AUC~$\uparrow$} & \textbf{AUC~$\uparrow$} & \textbf{AUC~$\uparrow$} &
\textbf{AUC~$\uparrow$} & \textbf{AUC~$\uparrow$} & \textbf{AUC~$\uparrow$} &
\textbf{AUC~$\uparrow$}&
\textbf{AUC~$\uparrow$}&
\textbf{AUC~$\uparrow$} \\
& &
$(n,c)$ &
$(n,c)$ &
$(n_1,n_2,n_3,n_4,c)$ &
$(n_1,n_2,n_3,n_4,c)$ &
$(n_1,n_2,n_3,n_4,c,\mu)$ &
$(n_1,n_2,n_3,n_4,c,\mu)$ &
$(d, \epsilon,\lambda, v1, v2, \theta)$&
$(n, c1, c3, \rho)$&
$(n_1,n_2,n_3,n_4,c,\mu,\gamma,\rho)$\\
\hline

\multirow{8}{*}{Chimpanzee\_vs\_Giantpanda}
&\multirow{2}{*}{5\%}&68.333&73.333&60.000&73.333&63.333&\textbf{80.000}&68.333&68.333&76.667\\
&&(15,$10^{-5}$)&(95,$10^{-5}$)&(25,21,105,1,$10^{2}$)&(35,11,25,1,$10^{1}$)&(35,7,35,1,$10^{4}$,0.0625)&($10^{1}$,13,95,1,$10^{3}$,0.5)&(1,$10^{-2}$,2,-2,0,2)&(35,$10^{-3}$,$10^{-4}$,1)&(35,21,105,1,$10^{4}$,$10^{-5}$,$10^{3}$,0.5)\\

&\multirow{2}{*}{10\%}&63.333&73.333&71.667&75.000&75.000&73.333&71.667&70.000&\textbf{83.333}\\
&&(65,$10^{-4}$)&(95,$10^{-5}$)&(45,3,5,1,$10^{1}$)&(50,13,75,1,$10^{2}$)&(40,13,105,1,$10^{3}$,0.03125)&(30,5,15,1,$10^{-4}$,0.03125)&($10^{1}$,$10^{-2}$,-2,-2,1,2)&(35,$10^{-3}$,$10^{-4}$,1)&(35,21,25,1,$10^{4}$,$10^{-5}$,$10^{-4}$,$10^{-2}$)\\

&\multirow{2}{*}{15\%}&70.000&65.000&60.000&73.333&70.000&\textbf{75.000}&71.667&71.667&\textbf{75.000}\\
&&(75,$10^{-5}$)&(105,$10^{-3}$)&(35,21,105,1,$10^{2}$)&(35,19,35,1,$10^{2}$)&(45,19,35,1,1,16)&(40,15,75,1,$10^{5}$,0.25)&(50,$10^{-2}$,2,-2,1,2)&(35,$10^{-3}$,$10^{-4}$,1)&(35,21,5,1,$10^{4}$,1,1,$10^{-2}$)\\

&\multirow{2}{*}{20\%}&70.000&76.667&73.333&68.333&63.333&71.667&71.667&68.333&\textbf{78.333}\\
&&(95,$10^{-3}$)&(95,$10^{-5}$)&($10^{1}$,13,105,1,$10^{2}$)&(20,21,65,1,$10^{2}$)&(25,19,35,1,1,16)&(20,21,75,1,1,16)&($10^{1}$,$10^{-2}$,-2,-2,1,2)&(35,$10^{-3}$,$10^{-4}$,1)&(25,7,35,1,$10^{1}$,$10^{-1}$,$10^{-5}$,2)\\
\hline

\multirow{8}{*}{Chimpanzee\_vs\_Hippopotamus}
&\multirow{2}{*}{5\%}&73.333&75.000&68.333&78.333&75.000&75.000&85.000&80.000&\textbf{88.333}\\
&&(25,$10^{-5}$)&(75,$10^{-3}$)&(40,19,15,1,$10^{3}$)&($10^{1}$,9,65,1,$10^{1}$)&(30,7,75,1,$10^{-5}$,4)&(30,9,85,1,$10^{-5}$,32)&(1,$10^{-2}$,-2,-2,1,2)&(5,$10^{2}$,1,$10^{3}$)&(5,19,5,1,1,$10^{-4}$,$10^{-5}$,1)\\

&\multirow{2}{*}{10\%}&71.667&76.667&66.667&76.667&76.667&78.333&\textbf{83.333}&76.667&81.667\\
&&(105,$10^{-5}$)&(75,$10^{-5}$)&(50,15,35,1,$10^{2}$)&($10^{1}$,9,65,1,$10^{1}$)&(25,9,25,1,$10^{-5}$,0.0625)&(30,7,85,1,$10^{-4}$,32)&(1,$10^{-2}$,-2,-2,1,2)&(5,$10^{2}$,1,$10^{3}$)&(5,19,85,1,$10^{3}$,$10^{-5}$,$10^{1}$,5)\\

&\multirow{2}{*}{15\%}&71.667&75.000&76.667&73.333&73.333&73.333&\textbf{81.667}&76.667&80.000\\
&&(65,$10^{-5}$)&(55,$10^{-4}$)&($10^{1}$,7,45,1,$10^{4}$)&(35,3,35,1,$10^{5}$)&($10^{1}$,7,45,1,$10^{-3}$,0.125)&(35,11,95,1,$10^{-3}$,32)&(1,$10^{-2}$,1,-2,1,2)&(5,$10^{2}$,1,$10^{3}$)&(5,19,55,1,$10^{4}$,$10^{-5}$,$10^{2}$,$10^{1}$)\\

&\multirow{2}{*}{20\%}&66.667&73.333&76.667&75.000&68.333&71.667&76.667&78.333&\textbf{85.000}\\
&&(75,$10^{-3}$)&(55,$10^{-2}$)&(5,21,55,1,$10^{2}$)&(20,9,5,1,$10^{5}$)&($10^{1}$,15,25,1,$10^{-5}$,0.03125)&(5,17,85,1,$10^{-5}$,4)&(50,$10^{-2}$,2,-2,2,2)&(5,$10^{2}$,1,$10^{3}$)&(5,19,5,1,1,$10^{-4}$,$10^{-5}$,1)\\
\hline

\multirow{8}{*}{Chimpanzee\_vs\_Humpbackwhale}
&\multirow{2}{*}{5\%}&78.333&91.667&90.000&91.667&90.000&91.667&91.667&93.333&\textbf{95.000}\\
&&(85,$10^{-3}$)&(55,1)&(15,15,25,1,$10^{4}$)&(15,9,65,1,$10^{3}$)&(45,9,5,1,$10^{-1}$,16)&(20,13,45,1,$10^{-5}$,8)&(1,$10^{-2}$,-2,1,1,2)&(35,$10^{-2}$,1,$10^{-1}$)&(50,13,85,1,$10^{3}$,$10^{3}$,$10^{-1}$,$10^{-1}$)\\

&\multirow{2}{*}{10\%}&85.000&91.667&85.000&91.667&90.000&91.667&\textbf{93.333}&90.000&\textbf{93.333}\\
&&(15,$10^{-3}$)&(55,$10^{-2}$)&(30,5,35,1,$10^{2}$)&(5,11,45,1,$10^{1}$)&(50,13,45,1,$10^{-1}$,16)&(20,5,45,1,1,16)&($10^{1}$,$10^{-2}$,-2,-2,1,2)&(35,$10^{-2}$,1,$10^{-1}$)&(50,13,55,1,$10^{5}$,$10^{-3}$,$10^{2}$,1)\\

&\multirow{2}{*}{15\%}&85.000&91.667&90.000&91.667&88.333&\textbf{91.667}&\textbf{91.667}&\textbf{91.667}&\textbf{91.667}\\
&&(45,$10^{-3}$)&(105,$10^{-5}$)&(5,13,35,1,$10^{3}$)&(15,13,25,1,$10^{3}$)&(5,13,35,1,$10^{5}$,0.5)&(35,9,105,1,$10^{-5}$,8)&(1,$10^{-2}$,-2,1,1,2)&(35,$10^{-2}$,1,$10^{-1}$)&(45,11,15,1,$10^{4}$,$10^{-2}$,$10^{1}$,$10^{1}$)\\

&\multirow{2}{*}{20\%}&86.667&88.333&88.333&93.333&90.000&90.000&93.333&90.000&\textbf{96.667}\\
&&(5,$10^{-2}$)&(105,$10^{-3}$)&($10^{1}$,11,45,1,$10^{1}$)&(15,15,25,1,$10^{5}$)&($10^{1}$,11,5,1,$10^{-3}$,32)&(35,9,55,1,$10^{5}$,4)&(1,$10^{-2}$,-2,-2,1,2)&(35,$10^{-2}$,1,$10^{-1}$)&(35,9,105,1,$10^{5}$,1,$10^{-5}$,$10^{-1}$)\\
\hline

\multirow{8}{*}{Chimpanzee\_vs\_Leopard}
&\multirow{2}{*}{5\%}&76.667&\textbf{81.667}&66.667&73.333&70.000&71.667&70.000&76.667&\textbf{81.667}\\
&&(65,$10^{-5}$)&(65,$10^{-3}$)&(30,5,25,1,$10^{5}$)&(30,19,85,1,$10^{2}$)&(30,5,45,1,$10^{2}$,0.125)&(20,15,55,1,$10^{5}$,1)&($10^{2}$,$10^{-2}$,-2,2,1,2)&(75,$10^{-4}$,$10^{1}$,1)&(40,13,105,1,$10^{3}$,$10^{1}$,$10^{-4}$,2)\\

&\multirow{2}{*}{10\%}&71.667&75.000&71.667&76.667&71.667&76.667&66.667&76.667&\textbf{86.667}\\
&&(55,$10^{-5}$)&(65,$10^{-2}$)&(45,11,65,1,$10^{4}$)&(20,21,45,1,$10^{4}$)&(45,11,65,1,$10^{4}$,0.0625)&(20,21,45,1,$10^{3}$,1)&(1,$10^{-2}$,-2,-2,0,2)&(75,$10^{-4}$,$10^{1}$,1)&(40,13,5,1,$10^{5}$,$10^{4}$,$10^{1}$,5)\\

&\multirow{2}{*}{15\%}&70.000&73.333&75.000&71.667&75.000&71.667&75.000&78.333&\textbf{85.000}\\
&&(115,$10^{-5}$)&(65,$10^{-5}$)&(40,3,75,1,$10^{5}$)&(25,21,5,1,$10^{3}$)&(40,3,65,1,$10^{-1}$,0.25)&(25,21,55,1,$10^{5}$,0.125)&(50,$10^{-2}$,-2,-2,1,2)&(75,$10^{-4}$,$10^{1}$,1)&(40,11,85,1,$10^{5}$,$10^{-2}$,1,$10^{-1}$)\\

&\multirow{2}{*}{20\%}&75.000&70.000&78.333&80.000&78.333&75.000&71.667&76.667&\textbf{81.667}\\
&&(5,$10^{-2}$)&(65,$10^{-5}$)&(40,17,55,1,$10^{4}$)&(25,21,85,1,$10^{3}$)&(50,7,65,1,$10^{1}$,0.25)&(25,21,85,1,$10^{5}$,0.0625)&(50,$10^{-2}$,-2,-2,1,2)&(75,$10^{-4}$,$10^{1}$,1)&(35,15,45,1,$10^{1}$,$10^{1}$,$10^{-3}$,$10^{-1}$)\\
\hline

\multirow{8}{*}{Chimpanzee\_vs\_Pig}
&\multirow{2}{*}{5\%}&76.667&73.333&71.667&66.667&68.333&71.667&73.333&63.333&\textbf{83.333}\\
&&(15,$10^{-5}$)&(95,$10^{-3}$)&($10^{1}$,3,65,1,$10^{4}$)&(45,7,15,1,$10^{2}$)&(25,15,55,1,$10^{-2}$,32)&(45,7,35,1,$10^{3}$,0.0625)&(1,$10^{-2}$,-2,1,0,2)&(35,$10^{-3}$,$10^{-4}$,1)&(45,21,25,1,$10^{2}$,$10^{-1}$,$10^{-3}$,$10^{-1}$)\\

&\multirow{2}{*}{10\%}&66.667&71.667&68.333&61.667&68.333&75.000&76.667&\textbf{76.667}&71.667\\
&&(25,$10^{-4}$)&(115,$10^{-3}$)&(50,11,35,1,$10^{2}$)&(45,7,85,1,$10^{5}$)&(15,9,65,1,$10^{-1}$,1)&(45,7,85,1,$10^{-1}$,1)&(1,$10^{-2}$,-2,1,0,2)&(5,$10^{1}$,$10^{2}$,$10^{1}$)&(45,7,45,1,1,$10^{-3}$,1,1)\\

&\multirow{2}{*}{15\%}&65.000&60.000&66.667&68.333&66.667&66.667&75.000&73.333&\textbf{78.333}\\
&&(75,$10^{-5}$)&(55,$10^{-4}$)&(45,1,75,1,$10^{3}$)&(15,17,55,1,$10^{3}$)&(20,17,45,1,1,0.25)&(15,17,55,1,$10^{5}$,0.5)&(1,$10^{-2}$,-2,-2,0,2)&(5,$10^{1}$,$10^{2}$,$10^{1}$)&(45,7,55,1,$10^{5}$,$10^{-3}$,$10^{4}$,0.5)\\

&\multirow{2}{*}{20\%}&56.667&65.000&68.333&66.667&70.000&65.000&73.333&73.333&\textbf{78.333}\\
&&(75,$10^{-3}$)&(55,$10^{-5}$)&($10^{1}$,19,25,1,$10^{5}$)&(45,17,25,1,$10^{2}$)&(20,13,25,1,$10^{-4}$,2)&(25,17,45,1,$10^{-2}$,32)&(1,$10^{-2}$,1,1,0,2)&(5,$10^{1}$,$10^{2}$,$10^{1}$)&(45,7,95,1,$10^{5}$,$10^{-5}$,1,$10^{-1}$)\\
\hline

\multirow{8}{*}{Chimpanzee\_vs\_Raccoon}
&\multirow{2}{*}{5\%}&70.000&63.333&73.333&66.667&76.667&65.000&70.000&68.333&\textbf{81.667}\\
&&(55,$10^{-4}$)&(55,$10^{-4}$)&($10^{1}$,11,105,1,$10^{1}$)&(20,21,85,1,$10^{5}$)&($10^{1}$,21,75,1,$10^{5}$,2)&(20,21,5,1,$10^{4}$,0.0625)&(1,$10^{-2}$,2,-2,2,2)&(5,$10^{2}$,1,$10^{3}$)&(35,17,15,1,$10^{2}$,$10^{2}$,1,2)\\

&\multirow{2}{*}{10\%}&68.333&70.000&75.000&70.000&76.667&70.000&71.667&75.000&\textbf{83.333}\\
&&(85,$10^{-4}$)&(55,$10^{-3}$)&(50,7,65,1,$10^{3}$)&(20,19,85,1,$10^{3}$)&(50,9,65,1,$10^{-3}$,0.0625)&(20,21,55,1,$10^{5}$,0.125)&(1,$10^{-2}$,-2,-2,-2,2)&(95,$10^{-2}$,$10^{2}$,$10^{3}$)&(35,21,25,1,$10^{5}$,$10^{5}$,$10^{3}$,5)\\

&\multirow{2}{*}{15\%}&60.000&71.667&71.667&\textbf{76.667}&70.000&71.667&70.000&75.000&65.000\\
&&(115,$10^{-4}$)&(55,$10^{-2}$)&($10^{1}$,19,105,1,$10^{3}$)&(20,21,45,1,$10^{4}$)&(40,7,15,1,$10^{1}$,0.25)&(20,9,105,1,$10^{-3}$,0.25)&(50,$10^{-2}$,-2,-2,0,2)&(95,$10^{-2}$,$10^{2}$,$10^{3}$)&(35,9,15,1,$10^{-5}$,1,$10^{-3}$,$10^{1}$)\\

&\multirow{2}{*}{20\%}&65.000&70.000&71.667&71.667&70.000&73.333&68.333&\textbf{83.333}&76.667\\
&&(35,$10^{-5}$)&(55,$10^{-5}$)&(20,19,45,1,$10^{5}$)&(20,5,95,1,$10^{2}$)&(20,17,75,1,$10^{5}$,0.0625)&(20,9,105,1,$10^{-3}$,0.25)&(1,$10^{-2}$,-2,-2,-2,2)&(95,$10^{-2}$,$10^{2}$,$10^{3}$)&(35,19,15,1,$10^{1}$,$10^{-2}$,$10^{-2}$,$10^{-2}$)\\
\hline

\multirow{8}{*}{Giantpanda\_vs\_Humpbackwhale}
&\multirow{2}{*}{5\%}&\textbf{93.333}&\textbf{93.333}&90.000&91.667&91.667&90.000&88.333&91.667&\textbf{93.333}\\
&&(25,$10^{-3}$)&(105,$10^{-3}$)&(20,11,95,1,$10^{2}$)&(20,7,85,1,$10^{3}$)&(20,11,65,1,$10^{-5}$,8)&(20,7,5,1,$10^{5}$,0.25)&(1,$10^{-2}$,-2,1,2,2)&(35,$10^{-2}$,1,$10^{-1}$)&(50,21,105,1,$10^{-1}$,$10^{-1}$,1,0.5)\\

&\multirow{2}{*}{10\%}&90.000&91.667&88.333&90.000&90.000&90.000&88.333&\textbf{93.333}&90.000\\
&&(25,$10^{-3}$)&(105,$10^{-3}$)&(20,5,35,1,$10^{5}$)&(20,7,85,1,$10^{3}$)&(45,21,25,1,1,4)&(40,11,35,1,$10^{-3}$,4)&(1,$10^{-2}$,-2,1,2,2)&(95,$10^{-1}$,$10^{3}$,$10^{2}$)&(15,13,55,1,1,$10^{-5}$,$10^{-2}$,5)\\

&\multirow{2}{*}{15\%}&91.667&\textbf{93.333}&90.000&91.667&91.667&91.667&90.000&\textbf{93.333}&\textbf{93.333}\\
&&(95,$10^{-3}$)&(105,$10^{-3}$)&(20,9,65,1,$10^{2}$)&(25,3,95,1,$10^{1}$)&(45,9,15,1,$10^{1}$,4)&(40,9,45,1,$10^{-4}$,8)&(1,$10^{-2}$,2,-2,2,2)&(95,$10^{-1}$,$10^{3}$,$10^{2}$)&(50,21,85,1,$10^{1}$,1,$10^{-3}$,0.5)\\

&\multirow{2}{*}{20\%}&90.000&\textbf{91.667}&\textbf{91.667}&90.000&\textbf{91.667}&88.333&90.000&90.000&\textbf{91.667}\\
&&(115,$10^{-3}$)&(115,$10^{-2}$)&(15,21,45,1,$10^{2}$)&(20,9,35,1,$10^{2}$)&(40,15,35,1,$10^{-2}$,8)&(40,9,45,1,$10^{-4}$,8)&(1,$10^{-2}$,2,-2,2,2)&(95,$10^{-1}$,$10^{3}$,$10^{2}$)&(15,21,105,1,$10^{3}$,$10^{2}$,$10^{-5}$,0.5)\\
\hline

\multirow{8}{*}{Giantpanda\_vs\_Pig}
&\multirow{2}{*}{5\%}&65.000&63.333&56.667&61.667&65.000&60.000&66.667&70.000&\textbf{73.333}\\
&&(35,$10^{-3}$)&(75,$10^{-3}$)&(35,13,75,1,$10^{2}$)&(35,17,25,1,$10^{2}$)&(30,19,85,1,$10^{2}$,8)&(30,11,65,1,$10^{-5}$,16)&(1,$10^{-2}$,-2,1,0,2)&(75,$10^{-4}$,$10^{1}$,1)&(50,21,85,1,$10^{1}$,1,$10^{-3}$,0.5)\\

&\multirow{2}{*}{10\%}&55.000&68.333&58.333&61.667&56.667&63.333&70.000&\textbf{76.667}&66.667\\
&&(65,$10^{-2}$)&(75,$10^{-2}$)&(50,11,35,1,$10^{2}$)&(35,17,5,1,$10^{2}$)&(20,19,95,1,$10^{4}$,8)&(45,9,95,1,$10^{4}$,2)&(1,$10^{-2}$,2,-2,0,2)&(45,$10^{-4}$,$10^{-3}$,$10^{-1}$)&(50,17,55,1,1,$10^{-3}$,1,1)\\

&\multirow{2}{*}{15\%}&58.333&63.333&60.000&65.000&56.667&70.000&66.667&\textbf{75.000}&\textbf{75.000}\\
&&(115,$10^{-4}$)&(65,$10^{-3}$)&(5,13,35,1,$10^{1}$)&(35,19,35,1,$10^{2}$)&(5,13,15,1,$10^{-3}$,32)&(40,7,95,1,$10^{-5}$,0.0625)&(50,$10^{-2}$,2,2,1,2)&(45,$10^{-4}$,$10^{-3}$,$10^{-1}$)&(50,21,35,1,$10^{1}$,$10^{3}$,$10^{-2}$,5)\\

&\multirow{2}{*}{20\%}&58.333&63.333&56.667&63.333&51.667&61.667&66.667&70.000&\textbf{71.667}\\
&&(85,$10^{-3}$)&(65,$10^{-3}$)&(50,11,105,1,$10^{2}$)&(35,17,5,1,$10^{2}$)&($10^{1}$,11,85,1,$10^{-3}$,2)&(30,5,35,1,$10^{-3}$,8)&(1,$10^{-2}$,2,-2,0,2)&(45,$10^{-4}$,$10^{-3}$,$10^{-1}$)&(50,21,85,1,$10^{-3}$,$10^{-4}$,$10^{-5}$,0.5)\\
\hline

\multirow{8}{*}{Leopard\_vs\_Hippopotamus}
&\multirow{2}{*}{5\%}&71.667&80.000&78.333&73.333&73.333&68.333&81.667&\textbf{83.333}&75.000\\
&&(45,$10^{-3}$)&(75,$10^{-2}$)&(35,5,45,1,$10^{1}$)&(45,7,105,1,$10^{2}$)&(5,11,35,1,$10^{1}$,4)&(40,11,45,1,$10^{-1}$,4)&(1,$10^{-2}$,-2,-2,0,2)&(5,$10^{1}$,$10^{2}$,$10^{1}$)&(50,9,45,1,$10^{5}$,$10^{-5}$,$10^{2}$,1)\\

&\multirow{2}{*}{10\%}&78.333&76.667&75.000&76.667&78.333&73.333&\textbf{80.000}&76.667&78.333\\
&&(55,$10^{-2}$)&(75,$10^{-3}$)&(50,15,35,1,$10^{2}$)&(45,17,15,1,$10^{2}$)&(45,19,35,1,1,16)&(40,13,95,1,$10^{1}$,8)&(1,$10^{-2}$,-2,-2,0,2)&(75,$10^{-2}$,$10^{-2}$,$10^{-2}$)&(5,19,5,1,1,$10^{-4}$,$10^{-5}$,1)\\

&\multirow{2}{*}{15\%}&61.667&\textbf{81.667}&78.333&80.000&73.333&71.667&\textbf{81.667}&76.667&\textbf{81.667}\\
&&(105,$10^{-2}$)&(55,$10^{2}$)&(40,3,95,1,$10^{2}$)&(45,13,15,1,$10^{2}$)&(35,17,65,1,$10^{-1}$,8)&(40,15,25,1,$10^{-5}$,8)&(1,$10^{-2}$,-2,-2,0,2)&(75,$10^{-2}$,$10^{-2}$,$10^{-2}$)&(50,9,45,1,$10^{5}$,$10^{-5}$,$10^{2}$,1)\\

&\multirow{2}{*}{20\%}&75.000&71.667&75.000&75.000&75.000&80.000&81.667&75.000&\textbf{83.333}\\
&&(65,$10^{-2}$)&(75,$10^{-4}$)&(5,9,5,1,$10^{1}$)&(30,7,65,1,$10^{1}$)&($10^{1}$,11,55,1,$10^{1}$,16)&(40,15,25,1,$10^{-5}$,8)&(1,$10^{-2}$,-2,1,0,2)&(75,$10^{-2}$,$10^{-2}$,$10^{-2}$)&(30,15,65,1,$10^{4}$,$10^{-1}$,$10^{2}$,$10^{-2}$)\\
\hline

\multirow{8}{*}{Leopard\_vs\_Persiancat}
&\multirow{2}{*}{5\%}&63.333&76.667&66.667&70.000&65.000&71.667&70.000&80.000&\textbf{83.333}\\
&&(75,$10^{-3}$)&(75,$10^{-3}$)&(45,5,75,1,$10^{1}$)&(20,3,85,1,$10^{1}$)&(35,11,85,1,$10^{1}$,0.03125)&(40,15,105,1,1,4)&(50,$10^{-2}$,1,-2,1,2)&(35,$10^{-3}$,$10^{-4}$,1)&(30,21,75,1,$10^{1}$,$10^{1}$,$10^{-3}$,0.5)\\

&\multirow{2}{*}{10\%}&60.000&76.667&71.667&76.667&70.000&66.667&76.667&\textbf{86.667}&85.000\\
&&(105,$10^{-3}$)&(65,$10^{-2}$)&(5,21,75,1,$10^{1}$)&(30,5,85,1,$10^{4}$)&(5,21,55,1,$10^{5}$,1)&(30,3,15,1,$10^{5}$,0.5)&(50,$10^{-2}$,-2,2,1,2)&(65,$10^{2}$,$10^{3}$,$10^{2}$)&(30,21,75,1,$10^{1}$,$10^{1}$,$10^{-3}$,0.5)\\

&\multirow{2}{*}{15\%}&60.000&78.333&70.000&56.667&68.333&73.333&71.667&\textbf{90.000}&81.667\\
&&(55,$10^{-3}$)&(65,$10^{-3}$)&(20,1,105,1,1)&(20,3,105,1,$10^{1}$)&(20,3,5,1,$10^{4}$,0.03125)&(40,7,25,1,$10^{-1}$,16)&(50,$10^{-2}$,-2,-2,1,2)&(65,$10^{2}$,$10^{3}$,$10^{2}$)&(30,21,75,1,$10^{-3}$,$10^{-4}$,$10^{-4}$,$10^{-1}$)\\

&\multirow{2}{*}{20\%}&61.667&73.333&68.333&75.000&43.333&61.667&75.000&\textbf{91.667}&83.333\\
&&(15,$10^{-5}$)&(95,$10^{-3}$)&($10^{1}$,21,85,1,$10^{2}$)&(20,5,75,1,$10^{1}$)&(30,11,75,1,$10^{-2}$,16)&(20,7,25,1,$10^{3}$,16)&($10^{2}$,$10^{-2}$,-2,2,1,2)&(65,$10^{2}$,$10^{3}$,$10^{2}$)&(50,19,95,1,$10^{-4}$,$10^{4}$,$10^{-4}$,$10^{1}$)\\
\hline

\multicolumn{10}{l}{Here, $^*$ denotes the proposed model.}
\end{tabular}
}
\end{adjustwidth}
\end{table*}

\begin{table*}[!htbp]
\ContinuedFloat
\centering
% \vspace*{-90pt}
\caption{\raggedright Continue.}
\begin{adjustwidth}{-2.8cm}{4.2cm}
\label{tab:awa_noise}
\setlength{\tabcolsep}{4pt} 
\renewcommand{\arraystretch}{1.3}
\resizebox{1.6\linewidth}{!}{%
\begin{tabular}{|c|c|c|c|c|c|c|c|c|c|c|c|}
\hline
 & & \textbf{ELM1~\cite{huang2006extreme}} & \textbf{ELM2~\cite{huang2006extreme}} & \textbf{BLS1~\cite{chen2017broad}} & \textbf{BLS2~\cite{chen2017broad}} & \textbf{IFBLS1~\cite{sajid2024intuitionistic}} & \textbf{IFBLS2~\cite{sajid2024intuitionistic}} &
 \textbf{MVLDM~\cite{hu2024multiview}} &
 \textbf{MVRVFL~\cite{quadir2024multiview}} &
\textbf{MVGIFBLS}$^*$\\\hline
\textbf{Dataset} & \textbf{Noise} & \textbf{AUC~$\uparrow$} & \textbf{AUC~$\uparrow$} & \textbf{AUC~$\uparrow$} &
\textbf{AUC~$\uparrow$} & \textbf{AUC~$\uparrow$} & \textbf{AUC~$\uparrow$} &
\textbf{AUC~$\uparrow$}&
\textbf{AUC~$\uparrow$}&
\textbf{AUC~$\uparrow$} \\
& &
$(n,c)$ &
$(n,c)$ &
$(n_1,n_2,n_3,n_4,c)$ &
$(n_1,n_2,n_3,n_4,c)$ &
$(n_1,n_2,n_3,n_4,c,\mu)$ &
$(n_1,n_2,n_3,n_4,c,\mu)$ &
$(d, \epsilon,\lambda, v1, v2, \theta)$&
$(n, c1, c3, \rho)$&
$(n_1,n_2,n_3,n_4,c,\mu,\gamma,\rho)$\\
\hline

\multirow{8}{*}{Persiancat\_vs\_Humpbackwhale}
&\multirow{2}{*}{5\%}&78.333&\textbf{86.667}&81.667&83.333&78.333&78.333&85.000&81.667&85.000\\
&&(85,$10^{-3}$)&(105,$10^{-5}$)&(45,3,5,1,$10^{5}$)&(15,7,35,1,$10^{3}$)&(20,5,95,1,$10^{-1}$,32)&(15,7,45,1,$10^{-5}$,0.0625)&($10^{2}$,$10^{-2}$,1,1,2,2)&(75,$10^{-4}$,$10^{1}$,1)&(40,21,85,1,$10^{4}$,$10^{4}$,$10^{1}$,$10^{1}$)\\

&\multirow{2}{*}{10\%}&\textbf{85.000}&81.667&80.000&81.667&80.000&80.000&81.667&83.333&\textbf{85.000}\\
&&(5,$10^{-5}$)&(95,$10^{-5}$)&($10^{1}$,19,15,1,$10^{5}$)&(15,7,45,1,$10^{3}$)&(50,17,105,1,$10^{-2}$,32)&(15,7,45,1,$10^{-5}$,0.0625)&(1,$10^{-2}$,-2,-2,-2,2)&(75,$10^{-1}$,$10^{2}$,$10^{4}$)&(40,19,45,1,$10^{-3}$,$10^{3}$,$10^{-1}$,1)\\

&\multirow{2}{*}{15\%}&81.667&83.333&80.000&85.000&73.333&80.000&83.333&85.000&\textbf{91.667}\\
&&(55,$10^{-3}$)&(95,$10^{-3}$)&(25,9,75,1,$10^{5}$)&(15,9,55,1,$10^{5}$)&(35,9,105,1,$10^{-2}$,32)&(15,5,5,1,$10^{-1}$,32)&(1,$10^{-2}$,2,2,1,2)&(75,$10^{-1}$,$10^{2}$,$10^{4}$)&(50,9,5,1,$10^{-1}$,$10^{-1}$,$10^{-4}$,$10^{-1}$)\\

&\multirow{2}{*}{20\%}&85.000&83.333&80.000&78.333&80.000&83.333&81.667&\textbf{88.333}&\textbf{88.333}\\
&&(35,$10^{-3}$)&(65,$10^{-5}$)&(15,21,55,1,$10^{1}$)&(15,7,45,1,$10^{2}$)&($10^{1}$,15,55,1,$10^{-1}$,32)&(50,15,55,1,$10^{-1}$,32)&(1,$10^{-2}$,-2,-2,-2,1)&(35,$10^{-2}$,1,$10^{-1}$)&(40,21,55,1,$10^{3}$,$10^{5}$,$10^{3}$,$10^{-1}$)\\
\hline

\multirow{8}{*}{Leopard\_vs\_Pig}
&\multirow{2}{*}{5\%}&63.333&65.000&53.333&63.333&61.667&63.333&66.667&65.000&\textbf{73.333}\\
&&(15,$10^{-3}$)&(75,$10^{-5}$)&(40,3,55,1,$10^{3}$)&(35,17,5,1,$10^{2}$)&(40,3,55,1,$10^{4}$,2)&(30,3,15,1,$10^{5}$,0.5)&(1,$10^{-2}$,-2,1,0,2)&(5,$10^{2}$,1,$10^{3}$)&(30,21,105,1,$10^{5}$,$10^{-5}$,$10^{-3}$,1)\\

&\multirow{2}{*}{10\%}&65.000&66.667&55.000&63.333&70.000&61.667&68.333&73.333&\textbf{75.000}\\
&&(95,$10^{-3}$)&(75,$10^{-3}$)&(50,11,85,1,$10^{1}$)&(35,19,75,1,$10^{2}$)&(40,3,85,1,$10^{2}$,2)&(20,13,105,1,1,32)&(1,$10^{-2}$,-2,1,0,2)&(25,1,$10^{-4}$,$10^{4}$)&(30,19,5,1,$10^{2}$,$10^{-5}$,$10^{-5}$,$10^{-2}$)\\

&\multirow{2}{*}{15\%}&50.000&56.667&58.333&58.333&56.667&58.333&65.000&68.333&\textbf{73.333}\\
&&(105,$10^{-2}$)&(55,$10^{-2}$)&(30,17,55,1,$10^{2}$)&(35,19,105,1,$10^{2}$)&(25,11,75,1,$10^{4}$,8)&(30,11,65,1,$10^{-5}$,8)&(1,$10^{-2}$,-2,1,0,2)&(25,1,$10^{-4}$,$10^{4}$)&(30,19,5,1,$10^{2}$,$10^{-5}$,$10^{-5}$,$10^{-2}$)\\

&\multirow{2}{*}{20\%}&61.667&63.333&63.333&68.333&61.667&63.333&68.333&70.000&\textbf{78.333}\\
&&(65,1)&(55,$10^{-2}$)&(45,13,105,1,$10^{1}$)&(35,19,75,1,$10^{2}$)&($10^{1}$,17,15,1,$10^{5}$,0.0625)&($10^{1}$,21,95,1,$10^{4}$,8)&(1,$10^{-2}$,-2,2,0,2)&(35,$10^{-3}$,$10^{-4}$,1)&(30,21,75,1,$10^{-3}$,$10^{-4}$,$10^{-4}$,$10^{-1}$)\\
\hline

\multirow{8}{*}{Persiancat\_vs\_Hippopotamus}
&\multirow{2}{*}{5\%}&70.000&76.667&75.000&78.333&63.333&70.000&78.333&75.000&\textbf{81.667}\\
&&(25,$10^{-2}$)&(75,$10^{-5}$)&(30,9,85,1,$10^{3}$)&(50,9,95,1,$10^{2}$)&(45,1,15,1,1,32)&(20,13,75,1,$10^{-5}$,8)&(50,$10^{-2}$,-2,-2,1,2)&(35,$10^{-2}$,1,$10^{-1}$)&(40,17,95,1,$10^{5}$,$10^{-1}$,$10^{1}$,$10^{-1}$)\\

&\multirow{2}{*}{10\%}&71.667&75.000&70.000&70.000&65.000&68.333&\textbf{83.333}&75.000&\textbf{83.333}\\
&&(55,$10^{-2}$)&(75,$10^{-3}$)&(15,3,65,1,$10^{1}$)&(15,11,95,1,$10^{4}$)&(35,7,95,1,$10^{3}$,0.125)&(20,17,95,1,$10^{1}$,8)&(1,$10^{-2}$,2,1,0,2)&(55,$10^{2}$,$10^{2}$,$10^{1}$)&(30,9,105,1,$10^{4}$,$10^{1}$,$10^{3}$,$10^{-2}$)\\

&\multirow{2}{*}{15\%}&56.667&78.333&75.000&78.333&55.000&75.000&\textbf{83.333}&76.667&81.667\\
&&(115,$10^{-2}$)&(35,$10^{-2}$)&(5,21,55,1,$10^{1}$)&($10^{1}$,19,105,1,$10^{3}$)&($10^{1}$,17,65,1,$10^{5}$,0.0625)&(50,5,85,1,$10^{5}$,16)&(1,$10^{-2}$,0,1,0,2)&(55,$10^{2}$,$10^{2}$,$10^{1}$)&(40,17,95,1,$10^{5}$,$10^{-1}$,$10^{1}$,$10^{-1}$)\\

&\multirow{2}{*}{20\%}&60.000&75.000&76.667&78.333&68.333&73.333&78.333&73.333&\textbf{88.333}\\
&&(105,$10^{-3}$)&(75,$10^{-2}$)&(20,11,95,1,$10^{2}$)&(50,9,105,1,$10^{2}$)&(20,7,45,1,$10^{-2}$,0.5)&(20,9,35,1,$10^{-1}$,16)&(50,$10^{-2}$,2,-2,1,2)&(5,$10^{2}$,1,$10^{3}$)&(15,21,15,1,1,$10^{-3}$,$10^{-1}$,5)\\

\multirow{8}{*}{Pig\_vs\_Hippopotamus}
&\multirow{2}{*}{5\%}&63.333&73.333&61.667&70.000&61.667&70.000&73.333&71.667&\textbf{76.667}\\
&&(75,$10^{-3}$)&(95,$10^{-5}$)&(50,21,65,1,$10^{3}$)&($10^{1}$,11,105,1,$10^{3}$)&(50,19,45,1,$10^{5}$,0.0625)&($10^{1}$,11,45,1,$10^{-5}$,0.25)&(50,$10^{-2}$,-2,-2,0,1)&(5,$10^{1}$,$10^{2}$,$10^{1}$)&(50,5,25,1,$10^{4}$,$10^{3}$,$10^{-2}$,1)\\

&\multirow{2}{*}{10\%}&71.667&68.333&63.333&71.667&65.000&71.667&75.000&70.000&\textbf{78.333}\\
&&(65,$10^{-5}$)&(105,$10^{-5}$)&(35,15,5,1,$10^{3}$)&(35,15,75,1,$10^{4}$)&(35,15,5,1,$10^{5}$,1)&(35,11,45,1,$10^{-2}$,1)&($10^{1}$,$10^{-2}$,-2,-2,2,2)&(55,$10^{-2}$,$10^{4}$,$10^{-3}$)&(50,3,25,1,$10^{5}$,$10^{-4}$,$10^{2}$,1)\\

&\multirow{2}{*}{15\%}&68.333&65.000&68.333&71.667&66.667&71.667&75.000&66.667&\textbf{80.000}\\
&&(5,$10^{-2}$)&(105,$10^{-5}$)&(35,19,95,1,$10^{3}$)&(35,19,75,1,$10^{3}$)&(35,19,55,1,$10^{5}$,2)&(50,5,25,1,$10^{3}$,0.03125)&($10^{2}$,$10^{-2}$,-2,2,0,2)&(55,$10^{-2}$,$10^{4}$,$10^{-3}$)&(50,3,35,1,$10^{4}$,$10^{3}$,$10^{-4}$,$10^{-2}$)\\

&\multirow{2}{*}{20\%}&66.667&70.000&63.333&66.667&63.333&61.667&76.667&53.333&\textbf{78.333}\\
&&(65,$10^{-1}$)&(105,$10^{-4}$)&(20,21,45,1,$10^{4}$)&(40,11,85,1,$10^{5}$)&(20,13,75,1,$10^{-5}$,0.25)&(40,11,85,1,$10^{-3}$,1)&(1,$10^{-2}$,-2,2,2,2)&(75,$10^{-4}$,$10^{1}$,1)&(50,3,35,1,$10^{4}$,$10^{3}$,$10^{-4}$,$10^{-2}$)\\
\hline

\multirow{8}{*}{Pig\_vs\_Raccon}
&\multirow{2}{*}{5\%}&65.000&65.000&70.000&70.000&66.667&65.000&65.000&75.000&\textbf{80.000}\\
&&(55,$10^{-3}$)&(115,$10^{-2}$)&(15,5,55,1,$10^{1}$)&(15,7,55,1,$10^{1}$)&(30,21,75,1,$10^{5}$,0.03125)&(30,21,35,1,$10^{4}$,0.5)&(1,$10^{-2}$,-2,-2,-2,2)&(95,$10^{-2}$,$10^{2}$,$10^{3}$)&(30,13,15,1,$10^{5}$,$10^{-5}$,$10^{-3}$,$10^{1}$)\\

&\multirow{2}{*}{10\%}&65.000&65.000&70.000&65.000&66.667&61.667&61.667&\textbf{80.000}&78.333\\
&&(15,$10^{-2}$)&(105,$10^{-3}$)&(15,17,35,1,$10^{1}$)&(15,7,75,1,$10^{1}$)&(45,19,75,1,$10^{3}$,0.125)&(30,21,35,1,$10^{4}$,0.5)&($10^{2}$,$10^{-2}$,2,1,-1,2)&(75,$10^{3}$,$10^{4}$,$10^{1}$)&(30,19,35,1,$10^{3}$,$10^{1}$,1,1)\\

&\multirow{2}{*}{15\%}&60.000&60.000&68.333&61.667&66.667&58.333&61.667&\textbf{78.333}&68.333\\
&&(5,$10^{-2}$)&(115,$10^{-2}$)&(5,9,95,1,$10^{2}$)&(30,21,65,1,$10^{3}$)&(5,9,35,1,$10^{-2}$,0.03125)&(30,9,95,1,$10^{1}$,1)&(1,$10^{-2}$,-2,-2,0,1)&(75,$10^{3}$,$10^{4}$,$10^{1}$)&(35,19,105,1,$10^{-3}$,$10^{1}$,$10^{-4}$,5)\\

&\multirow{2}{*}{20\%}&50.000&61.667&75.000&55.000&66.667&65.000&71.667&\textbf{85.000}&76.667\\
&&(5,$10^{-1}$)&(115,1)&(40,1,75,1,$10^{-2}$)&(15,7,75,1,$10^{1}$)&(15,19,35,1,$10^{-5}$,0.125)&(30,21,35,1,$10^{-5}$,0.125)&(1,$10^{-2}$,2,2,0,2)&(5,$10^{1}$,$10^{2}$,$10^{1}$)&(20,17,35,1,$10^{5}$,$10^{2}$,$10^{2}$,0.5)\\
\hline

\multirow{8}{*}{Raccoon\_vs\_Humpbackwhale}
&\multirow{2}{*}{5\%}&88.333&\textbf{90.000}&83.333&86.667&85.000&85.000&\textbf{90.000}&86.667&86.667\\
&&(55,$10^{-3}$)&(105,$10^{-3}$)&(40,7,35,1,$10^{2}$)&(40,15,5,1,$10^{2}$)&(25,17,35,1,$10^{-1}$,32)&(50,7,45,1,$10^{-1}$,32)&(1,$10^{-2}$,-2,2,0,2)&(95,$10^{-1}$,$10^{3}$,$10^{2}$)&(45,21,5,1,$10^{4}$,$10^{-4}$,$10^{3}$,1)\\

&\multirow{2}{*}{10\%}&86.667&90.000&83.333&90.000&85.000&86.667&\textbf{93.333}&88.333&88.333\\
&&(35,$10^{-3}$)&(105,$10^{-3}$)&(35,9,95,1,$10^{4}$)&(40,15,15,1,$10^{2}$)&(50,17,55,1,$10^{-2}$,32)&(50,7,45,1,$10^{-1}$,32)&(1,$10^{-2}$,-2,-2,-2,2)&(75,$10^{-2}$,1,$10^{4}$)&(45,21,45,1,$10^{-1}$,$10^{-3}$,$10^{-4}$,2)\\

&\multirow{2}{*}{15\%}&88.333&90.000&91.667&90.000&83.333&83.333&\textbf{93.333}&86.667&88.333\\
&&(45,$10^{-2}$)&(115,$10^{-3}$)&(45,3,15,1,$10^{5}$)&(25,7,95,1,$10^{2}$)&(45,3,85,1,$10^{-1}$,32)&(45,3,75,1,$10^{1}$,16)&(1,$10^{-2}$,-2,-2,0,2)&(75,$10^{-2}$,1,$10^{4}$)&(40,17,15,1,1,1,$10^{-2}$,5)\\

&\multirow{2}{*}{20\%}&88.333&81.667&83.333&85.000&83.333&90.000&\textbf{93.333}&85.000&91.667\\
&&(105,$10^{-3}$)&(105,$10^{-3}$)&(5,9,105,1,$10^{1}$)&(40,15,55,1,$10^{2}$)&(35,1,5,1,1,0.03125)&(40,17,105,1,1,16)&(1,$10^{-2}$,2,-2,0,2)&(95,$10^{-2}$,$10^{2}$,$10^{3}$)&(15,21,55,1,$10^{5}$,$10^{4}$,$10^{4}$,1)\\
\hline
\multicolumn{2}{|l|}{\textbf{Average AUC~$\uparrow$}} & 70.860 & 75.180 & 72.970 & 74.690 & 71.880 & 73.800 & 77.190 & 78.330 & \textbf{81.900} \\ \hline
\multicolumn{2}{|l|}{\textbf{Average Rank~$\downarrow$}} & 6.81 & 4.96 & 6.26 & 5.05 & 6.56 & 5.91 & 3.88 & 3.66 & \textbf{1.91} \\ \hline
\hline
\multicolumn{10}{l}{Here, $^*$ denotes the proposed model.}
\end{tabular}
}
\end{adjustwidth}
\end{table*}

\subsection{Statistical Analysis}
We use statistical tests, including Friedman, Wilcoxon signed-rank, and win–tie–loss comparisons, on the UCI and KEEL datasets to evaluate the performance of MVGIFBLS against baseline models.

\subsubsection{Friedman Analysis}
The Friedman test \cite{friedman1940comparison} is a non-parametric method used to detect significant performance differences among multiple algorithms across several datasets.Unlike ANOVA, it does not require the data to follow a normal distribution. It compares the rankings of algorithms instead of their raw performance values, which makes it suitable for this analysis. The null hypothesis of the Friedman test states that all models perform equally and have similar average ranks. In our study, we evaluated nine models: ELM1, ELM2, BLS1, BLS2, IFBLS1, IFBLS2, MVLDM, MVRVFL, and the proposed MVGIFBLS model over 13 datasets. The average ranks of these models are 4.12, 5.08, 5.73, 5.42, 5.77, 6.31, 5.69, 4.00, and 2.88, respectively. 

The Friedman chi-square statistic is computed as:  

\begin{equation}
\chi_F^2 = \frac{12N}{w(w+1)} \left( \sum_{j=1}^w I_j^2 - \frac{w(w+1)^2}{4} \right),
\end{equation}

where $N$ is the number of datasets, $w$ is the number of models, and $I_j$ represents the average rank of the $j^{th}$ model. To adjust for finite sample sizes, the Friedman statistic is further transformed into an F-distribution using:  

\begin{equation}
F_F = \frac{(N-1)\chi_F^2}{N(w-1) - \chi_F^2}.
\end{equation}

The Friedman statistic is computed as $\chi_F^2 = 16.9340$, indicating that the null hypothesis of equal performance can be rejected. The adjusted Friedman F-statistic is obtained as $F_F = 2.3340$. 

\begin{figure}[h!]
\centering
\includegraphics[width=0.9\linewidth]{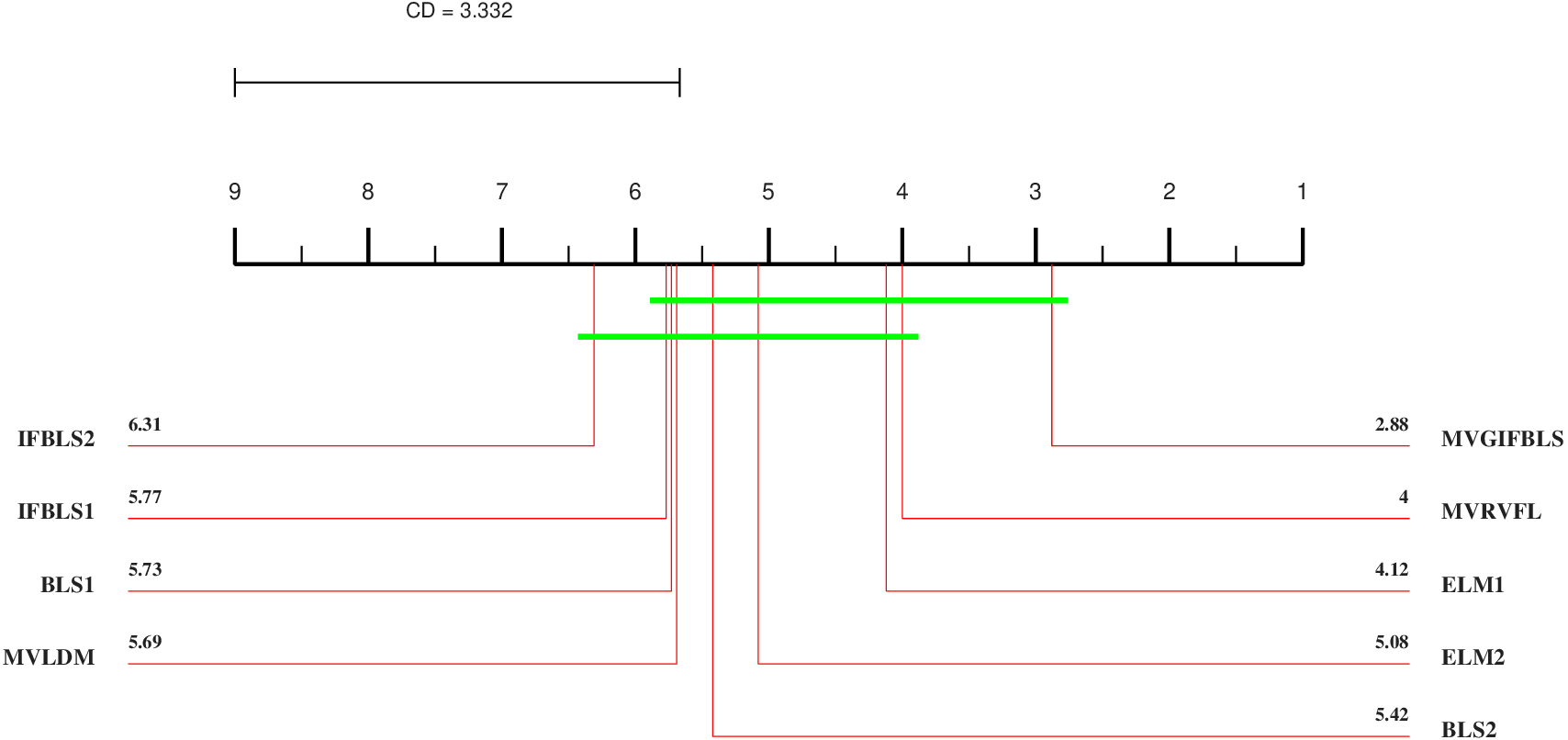}
\caption{Critical difference (CD) diagram based on the Nemenyi test at the 5\% significance level.}
\label{fig:cd_diagram}
\end{figure}

To identify which models differ, the Nemenyi post-hoc test was applied. The critical difference (CD) is given by:  

\begin{equation}
CD = q_\alpha \sqrt{\frac{w(w+1)}{6N}},
\end{equation}

where $q_\alpha$ is the critical value based on the studentized range statistic. At the 5\% level of significance, $q_\alpha = 3.1020$, and the corresponding critical difference is $CD = 3.3300$.
Pairwise comparisons based on the Nemenyi test ( Fig.~\ref{fig:cd_diagram}) show that the proposed MVGIFBLS model achieves a statistically significant improvement over IFBLS2, as the difference in their average ranks exceeds the critical difference (CD = 3.332). However, the differences between MVGIFBLS and the remaining models do not exceed the CD, indicating that their performances are statistically comparable. Overall, the proposed MVGIFBLS model attains the average rank of 2.88, demonstrating strong and consistent performance across the datasets.

\subsubsection{Wilcoxon Analysis}
The Wilcoxon signed-rank test \cite{demvsar2006statistical} is a non-parametric method used to compare the performance of two models across multiple datasets. It does not assume that the data follow a normal distribution. Instead, it uses the relative ranking of performance differences between paired samples. This makes it suitable for evaluating learning algorithms on multiple datasets. Given two models, let $d^f_i$ denote the difference in performance (AUC) between the proposed MVGIFBLS model and the baseline on the $i^{th}$ dataset. The Wilcoxon test calculates two rank-based statistics: 

\begin{equation}
R^+ = \sum_{d^f_i > 0} \text{rank}(d^f_i) + \frac{1}{2}\sum_{d^f_i = 0} \text{rank}(d^f_i),
\label{eq:wilcoxonRplus}
\end{equation}

\begin{equation}
R^- = \sum_{d^f_i < 0} \text{rank}(d^f_i) + \frac{1}{2}\sum_{d^f_i = 0} \text{rank}(d^f_i),
\label{eq:wilcoxonRminus}
\end{equation}

were, $R^+$ is the sum of the ranks of datasets where the proposed MVGIFBLS model outperforms the baseline, while $R^-$ is the sum of the ranks of datasets where the baseline outperforms the proposed MVGIFBLS model. The test statistic is defined as:
\begin{equation}
T = \min(R^+, R^-).
\end{equation}
The corresponding $p$-value is then computed to evaluate statistical significance. If $p < 0.05$ (5\% significance level), we reject the null hypothesis, indicating that the observed difference is statistically significant.
Table \ref{tab:my_label1} reports the results of the Wilcoxon signed-rank test for the proposed MVGIFBLS model against eight baseline methods: ELM1, ELM2, BLS1, BLS2, 
IFBLS1, IFBLS2, MVLDM, and MVRVFL. The values of $R^+$, $R^-$, $p$ values, and hypothesis decisions are presented.
\begin{table*}[htbp]
\caption{The Wilcoxon signed-rank test conducted to compare the baseline models with the proposed MVGIFBLS model.}
\vspace{-\baselineskip} 
    \begin{center}
    \begin{tabular}{|c|c|c|c|c|}
        \hline 
       Model & $R+$ & $R-$ & p-value & Hypothesis (0.05) \\
        \hline    
        ELM1 \cite{huang2006extreme}  & 58.00 & 20.00 & .0671 & NR \\
         \hline  
        ELM2 \cite{huang2006extreme}  & 66.50 & 11.50 & .0151 & R \\
         \hline  
        BLS1 \cite{chen2017broad} & 84.50 & 6.50 &  .0032 & R \\
         \hline  
        BLS2 \cite{chen2017broad}  & 74.00 & 4.00 & .0029 & R \\
         \hline  
        IFBLS1 \cite{sajid2024intuitionistic}  & 78.50 & 12.50 & .0104 & R \\
         \hline 
        IFBLS2 \cite{sajid2024intuitionistic} & 65.00 & 1.00 &  .0021 & R \\\hline 
        MVLDM \cite{hu2024multiview} & 79.00 & 12.00 &  .0094 & R \\\hline 
        MVRVFL \cite{quadir2024multiview} & 60.50 & 30.50 &  .1456 & NR\ \\
        \hline
    \end{tabular}
 \label{tab:my_label1}
 \end{center}
\end{table*}
From Table \ref{tab:my_label1}, the proposed MVGIFBLS model achieves statistically significant improvements over ELM2, BLS1, BLS2, IFBLS1, IFBLS2, and MVLDM, as their $p$-values are below 0.05, resulting in the rejection of the null hypothesis. For example, when compared with BLS1, the Wilcoxon signed-rank test produces $R^+ = 84.50$, $R^- = 6.50$, and a $p$-value of 0.0032, confirming that the observed performance difference is statistically significant.
However, for ELM1 ($p = 0.0671$) and MVRVFL ($p = 0.1456$), the $p$-values are greater than 0.05, so the null hypothesis is not rejected. This indicates that the performance difference between MVGIFBLS and these two models is not statistically significant. Overall, the results show that the proposed MVGIFBLS model performs better than six out of eight baseline models, demonstrating its effectiveness and robustness.

\subsubsection{Pairwise Win-Tie-Loss}

The Win–Tie–Loss test \cite{demvsar2006statistical} compares two models across multiple datasets by counting the number of wins, ties, and losses.A win indicates that the row model performs better, a tie indicates equal performance, and a loss indicates that the row model performs worse than the column model. We use this test to compare the proposed MVGIFBLS model with the baseline models, including ELM1, ELM2, BLS1, BLS2, IFBLS1, IFBLS2, MVLDM, and MVRVFL. The results are reported as [wins, ties, losses]. 

\begin{table*}[h!]
\caption{Win--Tie--Loss comparison of the proposed MVGIFBLS model with baseline models.}
\label{tab:my_label2}
\centering
\setlength{\tabcolsep}{1.2pt}
\renewcommand{\arraystretch}{1.3}
\resizebox{\linewidth}{!}{%
\begin{tabular}{|c|c|c|c|c|c|c|c|c|}
\hline
Models & ELM1 & ELM2 & BLS1 & BLS2 & IFBLS1 & IFBLS2 & MVLDM & MVRVFL \\
\hline
ELM2 \cite{huang2006extreme} & $[6,0,7]$ & & & & & & &\\
\hline
BLS1 \cite{chen2017broad} & $[4,0,9]$ & $[5,0,8]$ & & & & & &\\
\hline
BLS2 \cite{chen2017broad} & $[4,0,9]$ & $[6,2,5]$ & $[6,1,6]$ & & & & &\\
\hline
IFBLS1 \cite{sajid2024intuitionistic} & $[4,0,9]$ & $[5,0,8]$ & $[6,1,6]$ & $[5,0,8]$ & & & &\\
\hline
IFBLS2 \cite{sajid2024intuitionistic} & $[2,0,11]$ & $[3,1,9]$ & $[4,1,8]$ & $[6,2,5]$ & $[6,0,7]$ & & &\\
\hline
MVLDM \cite{hu2024multiview} & $[7,0,6]$ & $[6,1,6]$ & $[6,2,5]$ & $[4,0,9]$ & $[5,1,7]$ & $[6,0,7]$ & &\\
\hline
MVRVFL \cite{quadir2024multiview} & $[6,0,7]$ & $[8,1,4]$ & $[9,0,4]$ & $[8,1,4]$ & $[9,0,4]$ & $[10,0,3]$ & $[8,0,5]$ &\\
\hline
MVGIFBLS$^*$ & $[7,1,5]$ & $[10,1,2]$ & $[11,0,2]$ & $[11,1,1]$ & $[10,0,3]$ & $[10,2,1]$ & $[11,0,2]$ & $[7,0,6]$\\
\hline
\multicolumn{6}{l}{Here, $^*$ denotes the proposed MVGIFBLS model.}
\end{tabular}
}
\end{table*}

From Table \ref{tab:my_label2}, the proposed MVGIFBLS model shows strong performance against all baseline models. It achieves more wins than losses against ELM1, ELM2, BLS1, and BLS2, indicating clear improvement in most cases. It also performs better than IFBLS1, IFBLS2, and MVLDM with consistently higher wins. Overall, these results show that MVGIFBLS outperforms most baseline models across the datasets.
When compared with MVRVFL, the results are [7, 0, 6], which indicates competitive performance with a similar number of wins and losses. Overall, MVGIFBLS achieves more wins than losses in almost all comparisons, demonstrating its effectiveness and consistency across different datasets.

\subsection{Hyperparameter Sensitivity Analysis on UCI and KEEL Datasets}
\begin{figure}[H]
  \centering
  % First row
  \begin{minipage}[b]{0.45\textwidth}
    \centering
    \includegraphics[width=\textwidth]{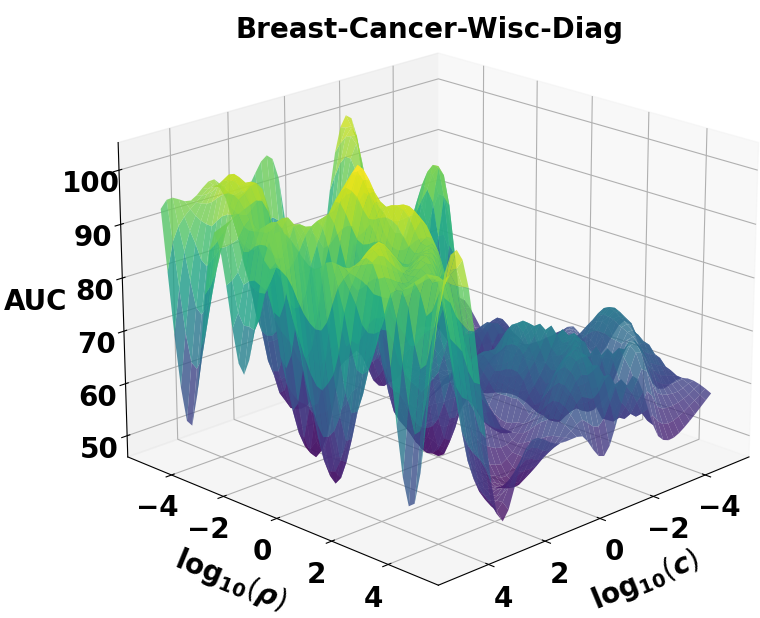}
  \end{minipage} \hfill
  \begin{minipage}[b]{0.45\textwidth}
    \centering
    \includegraphics[width=\textwidth]{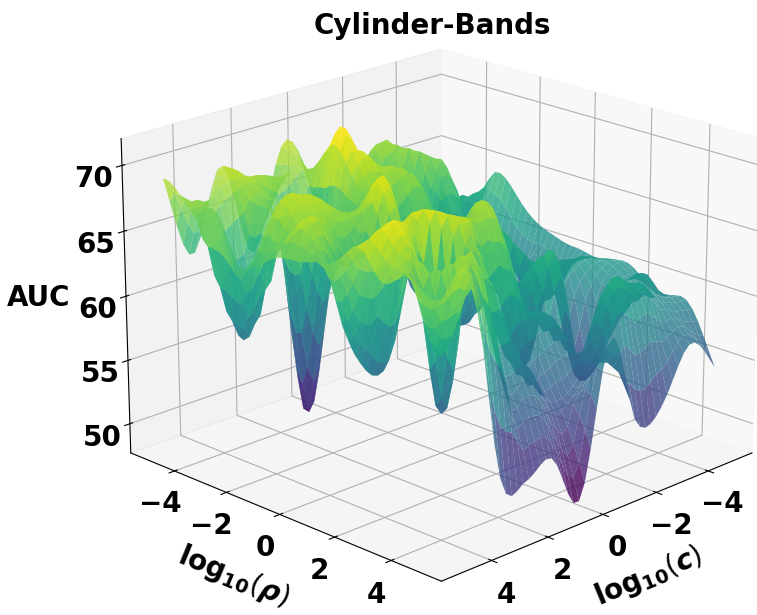}
  \end{minipage}\\[2ex]

  % \vspace{0.5cm} % Add space between rows

  % Second row
  \begin{minipage}[b]{0.45\textwidth}
    \centering
    \includegraphics[width=\textwidth]{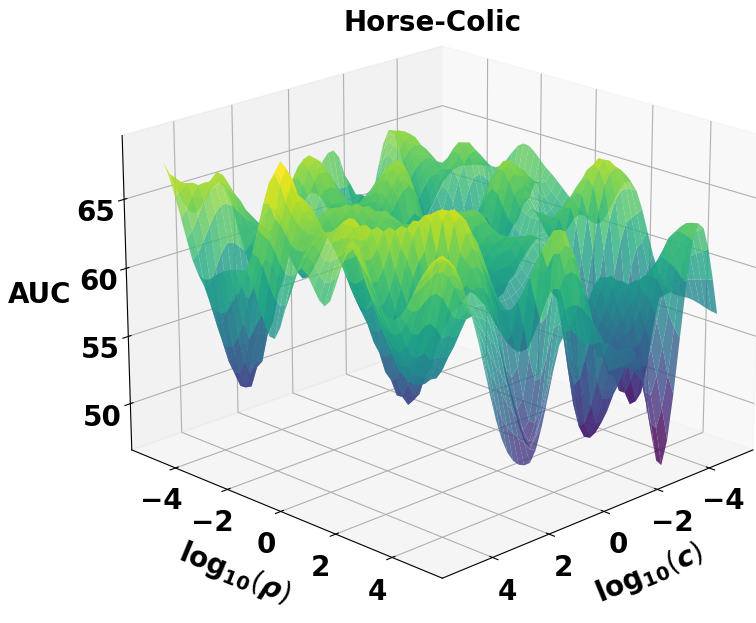}
  \end{minipage} \hfill
  \begin{minipage}[b]{0.45\textwidth}
    \centering
    \includegraphics[width=\textwidth]{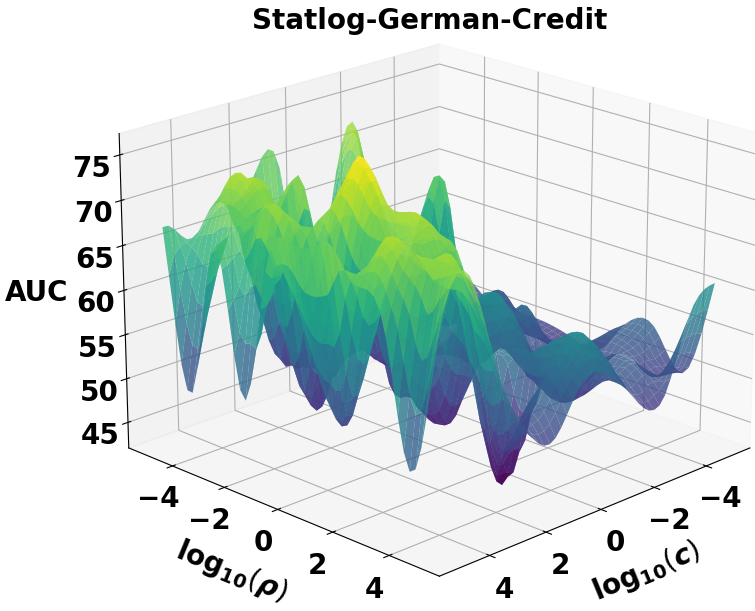}
  \end{minipage}

\caption{Sensitivity analysis of the proposed MVGIFBLS model with different values of $c$ and $\rho$ in terms of AUC performance on the UCI and KEEL datasets.}
  \label{fig:sensitivity}
\end{figure}

We conduct a hyperparameter sensitivity analysis to examine the effect of the parameters $c$ and $\rho$ on the performance of the proposed MVGIFBLS model. Fig.~\ref{fig:sensitivity} presents the 3D surface plots for four benchmark datasets: Breast-cancer-wisc-diag, Cylinder-bands, Horse-colic, and Statlog-German-Credit. The horizontal axes represent the $\log_{10}(c)$ and $\log_{10}(\rho)$ values, while the vertical axis represents the AUC score.From the plots, we observe that model performance varies with different values of $c$ and $\rho$. For the Breast-cancer-wisc-diag dataset, the model achieves high AUC values over several parameter regions, which indicates stable performance. For the Cylinder-bands and Horse-colic datasets, the AUC varies moderately with changes in parameter values, showing that proper parameter selection improves classification performance. In the Statlog-German-Credit dataset, the model obtains better results in specific regions of the parameter space, while extreme values of $c$ and $\rho$ reduce the performance.
Overall, the results show that the proposed MVGIFBLS model maintains good performance across a wide range of parameter values. The analysis also demonstrates that selecting suitable values of $c$ and $\rho$ helps improve the robustness and classification ability of the model.

\subsection{Hyperparameter Sensitivity Analysis on AwA Datasets}
\begin{figure}[H]
  \centering
  % First row
  \begin{minipage}[b]{0.45\textwidth}
    \centering
    \includegraphics[width=\textwidth]{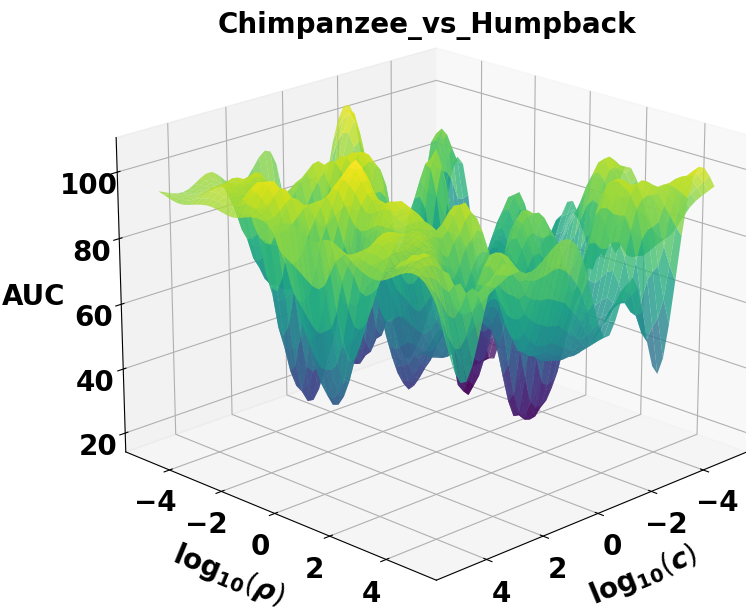}
  \end{minipage} \hfill
  \begin{minipage}[b]{0.45\textwidth}
    \centering
    \includegraphics[width=\textwidth]{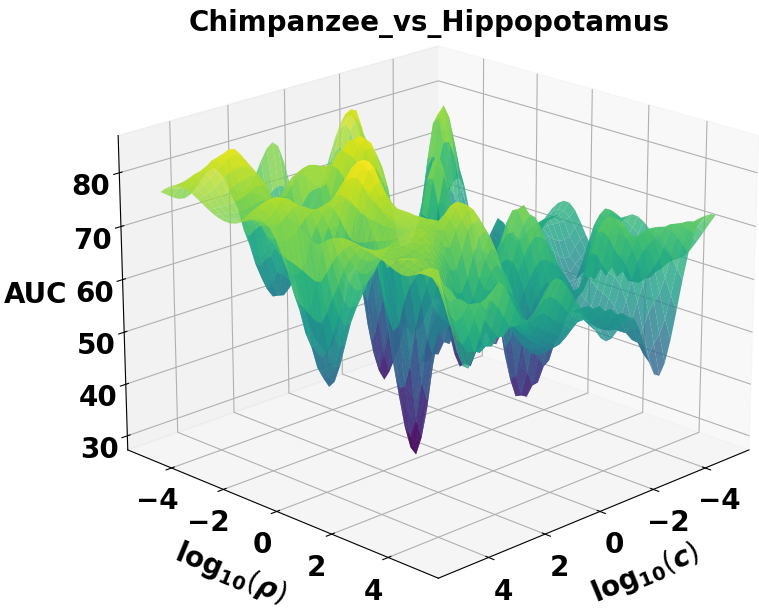}
  \end{minipage}\\[2ex]

  % \vspace{0.5cm} % Add space between rows

  % Second row
  \begin{minipage}[b]{0.45\textwidth}
    \centering
    \includegraphics[width=\textwidth]{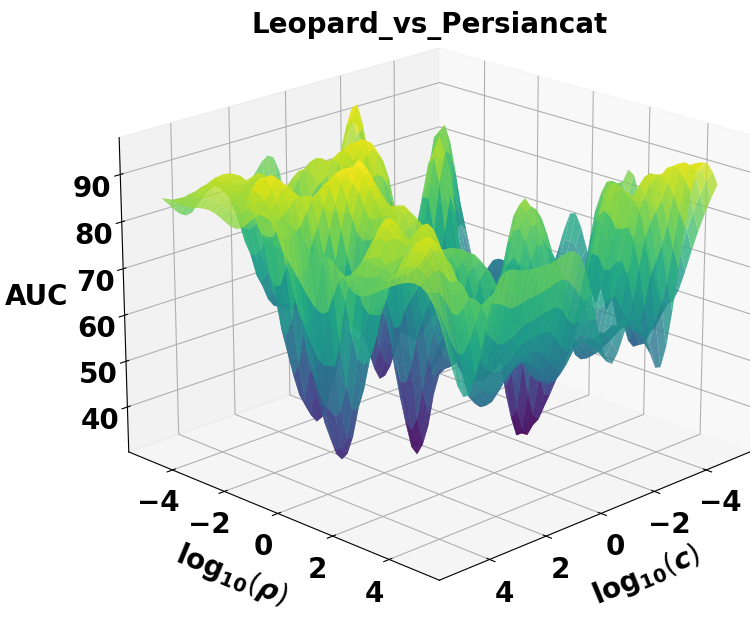}
  \end{minipage} \hfill
  \begin{minipage}[b]{0.45\textwidth}
    \centering
    \includegraphics[width=\textwidth]{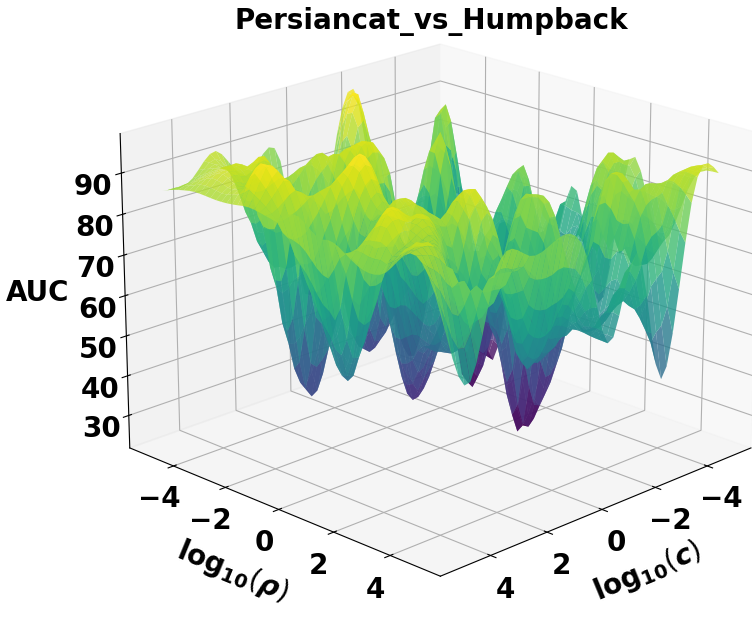}
  \end{minipage}

\caption{Sensitivity analysis of the proposed MVGIFBLS model with different values of $c$ and $\rho$ in terms of AUC performance on the AwA datasets.}
  \label{fig:awa_sensitivity}
\end{figure}

We evaluate the effect of different values of $c$ and $\rho$ on the performance of the proposed MVGIFBLS model for the AwA datasets. Fig.~\ref{fig:awa_sensitivity} shows the 3D surface plots for the four datasets Chimpanzee\_vs\_Humpback, Chimpanzee\_vs\_Hippopotamus, Leopard\_vs\_Persiancat, and Persiancat\_vs\_Humpback. The horizontal axes represent the $\log_{10}(c)$ and $\log_{10}(\rho)$ values, while the vertical axis represents the AUC score. From the plots, we observe that the AUC values change with different combinations of $c$ and $\rho$. 
For the Chimpanzee\_vs\_Humpback dataset, the model achieves high AUC values across several parameter regions, which indicates stable classification performance. In the Chimpanzee\_vs\_Hippopotamus dataset, the AUC varies considerably, showing that suitable parameter values are important for better performance. Similar behavior is observed for the Leopard\_vs\_Persiancat and Persiancat\_vs\_Humpback datasets, where moderate parameter values generally provide better results, while extreme values reduce the AUC performance. Overall, the results show that the proposed MVGIFBLS model achieves stable and reliable performance over a wide range of parameter values. The analysis also highlights the importance of selecting appropriate values of $c$ and $\rho$ to obtain better classification results on the AwA datasets.

\subsection{Ablation Study on the AwA Datasets}

To understand the contribution of each component in the proposed MVGIFBLS framework, we performed an ablation study on the AwA datasets. The results are reported in Table~\ref{tab:ablation}. We evaluated six variants: (i) the proposed MVGIFBLS model, (ii) MVGIFBLS without Graph Embedding (w/o GE), (iii) MVGIFBLS without Intuitionistic Fuzzy weighting (w/o IF), (iv) MVGIFBLS without Cross-View Consistency (w/o CV), (v) using only View-E, and (vi) using only View-F.

The proposed MVGIFBLS model achieves the highest average AUC of 82.71\%. When Graph Embedding is removed, the average AUC decreases slightly to 82.50\%. This result shows that Graph Embedding helps the model preserve the geometric relationships among samples and contributes to improved discrimination between classes. Although the reduction is small, the proposed model still provides the highest overall performance.

\begin{table*}[t]
\centering
\vspace*{-70pt}
\caption{Ablation Study Results (AUC \%) on AwA Datasets}
\label{tab:ablation}
\resizebox{\textwidth}{!}{
\begin{tabular}{lcccccc}
\hline
\textbf{Dataset} 
    & \textbf{MVGIFBLS $^*$} 
    & \textbf{w/o GE} 
    & \textbf{w/o IF} 
    & \textbf{w/o CV} 
    & \textbf{View-E Only} 
    & \textbf{View-F Only} \\
\hline
Chimpanzee\_vs.\_Giant Panda        & \textbf{76.67} & 75.00 & 66.67 & \textbf{76.67} & 75.00 & \textbf{76.67} \\
Chimpanzee\_vs.\_Hippopotamus       & \textbf{85.00} & \textbf{85.00} & 73.33 & \textbf{85.00} & 83.33 & 76.67 \\
Chimpanzee\_vs.\_Humpback Whale     & \textbf{96.67} & \textbf{96.67} & 88.33 & \textbf{96.67} & 91.67 & 93.33 \\
Chimpanzee\_vs.\_Leopard            & 80.00 & \textbf{85.00} & 71.67 & 80.00 & 75.00 & 75.00 \\
Chimpanzee\_vs.\_Pig                & \textbf{80.00} & \textbf{80.00} & 73.33 & \textbf{80.00} & 76.67 & 75.00 \\
Chimpanzee\_vs.\_Raccoon            & 81.67 & 80.00 & \textbf{83.33} & 78.33 & 71.67 & 76.67 \\
Giant Panda\_vs.\_Humpback Whale    & \textbf{93.33} & \textbf{93.33} & 88.33 & \textbf{93.33} & 91.67 & 73.33 \\
Giant Panda\_vs.\_Pig               & \textbf{76.67} & \textbf{76.67} & 75.00 & \textbf{76.67} & 71.67 & 60.00 \\
Leopard\_vs.\_Hippopotamus          & \textbf{80.00} & \textbf{80.00} & 65.00 & \textbf{80.00} & 48.33 & 75.00 \\
Leopard\_vs.\_Persian Cat           & \textbf{83.33} & \textbf{83.33} & 81.67 & \textbf{83.33} & \textbf{83.33} & 68.33 \\
Leopard\_vs.\_Pig                   & \textbf{73.33} & \textbf{73.33} & 65.00 & \textbf{73.33} & \textbf{73.33} & 63.33 \\
Persian Cat\_vs.\_Hippopotamus      & \textbf{80.00} & 78.33 & 71.67 & \textbf{80.00} & \textbf{80.00} & 70.00 \\
Persian Cat\_vs.\_Humpback Whale    & \textbf{90.00} & \textbf{90.00} & \textbf{90.00} & 88.33 & 86.67 & 81.67 \\
Pig\_vs.\_Hippopotamus              & \textbf{75.00} & \textbf{75.00} & 71.67 & \textbf{75.00} & 48.33 & \textbf{75.00} \\
Pig\_vs.\_Raccoon                   & \textbf{81.67} & 78.33 & 75.00 & \textbf{81.67} & 75.00 & 71.67 \\
Raccoon\_vs.\_Humpback Whale        & \textbf{90.00} & \textbf{90.00} & 88.33 & \textbf{90.00} & 83.33 & 80.00 \\
\hline
\textbf{Average}                   & \textbf{82.71} & 82.50 & 76.77 & 82.40 & 75.94 & 74.48 \\
\hline
 \multicolumn{6}{l}{Here, $^*$ denotes the proposed model.}
\end{tabular}
}
\end{table*}

The largest performance drop is observed when the Intuitionistic Fuzzy component is removed. The average AUC decreases from 82.71\% to 76.77\%, resulting in a reduction of 5.94 percentage points. This finding indicates that the intuitionistic fuzzy weighting mechanism plays a major role in the proposed framework. Assigning adaptive weights to training samples, it reduces the influence of noisy samples and outliers and improves the robustness of the classifier.

When the Cross-View Consistency term is removed, the average AUC decreases slightly from 82.71\% to 82.40\%. This result demonstrates that encouraging agreement between the two views helps improve the learning process. The consistency constraint allows both views to share complementary information and produces more stable predictions.

The single-view experiments further highlight the importance of multi-view learning. Using only View-E achieves an average AUC of 75.94\%, while using only View-F achieves 74.48\%. Both results are lower than the performance of the proposed MVGIFBLS model. This improvement confirms that combining information from both views provides richer feature representations and better classification performance than relying on a single view alone.

Overall, the ablation study demonstrates that each component contributes positively to the final performance. Among all components, the Intuitionistic Fuzzy mechanism provides the largest improvement, while Graph Embedding and Cross-View Consistency further enhance the discriminative ability and stability of the framework. The Highest performance of the proposed MVGIFBLS model confirms that the integration of Multi-View Learning, Graph Embedding, and Intuitionistic Fuzzy theory is effective for AwA classification tasks.

\section{Conclusion}
\label{section5}

In this study, we proposed the MVGIFBLS framework by integrating multi-view learning, graph embedding, and intuitionistic fuzzy theory into the Broad Learning System. The proposed framework improves the generalization and classification capability of BLS by effectively learning complementary information from multiple data views. Graph embedding preserves the structural relationships among samples, while intuitionistic fuzzy theory improves robustness to noise and outliers. The proposed MVGIFBLS model was evaluated on UCI, KEEL, and AwA benchmark datasets and compared with ELM1, ELM2, BLS1, BLS2, IFBLS1, IFBLS2, MVLDM, and MVRVFL. The results show that the proposed MVGIFBLS model achieves the highest average AUC and the lowest average rank among all compared methods. In terms of relative improvement, the proposed MVGIFBLS model achieves AUC gains ranging from 2.35\% to 6.70\% on the UCI and KEEL datasets and from 5.84\% to 12.05\% on the AwA dataset compared with the baseline methods.
The Gaussian feature noise experiments further demonstrate that the proposed framework maintains stable classification performance under different noise levels, confirming its robustness to corrupted input features. In addition, the ablation study verifies that graph embedding, intuitionistic fuzzy weighting, cross-view consistency, and multi-view feature fusion each contribute to the overall performance of the proposed MVGIFBLS framework. Statistical analyses, including average AUC, ranking schemes, the Friedman test, the Wilcoxon signed-rank test, and the win--tie--loss analysis, further confirm that the proposed MVGIFBLS model outperforms the baseline methods. The current study is limited to binary classification problems. In future work, we plan to extend the proposed MVGIFBLS framework to multiclass classification tasks and datasets containing more than two views. We also aim to investigate adaptive graph construction strategies and computationally efficient optimization techniques to improve scalability for large-scale applications.

\section*{Declarations}

\textbf{Funding:} This research received no external funding.

\noindent
\textbf{Conflicts of interest:} Authors confirm that there is no conflict of interest with anyone involved

\bibliography{bibfile}

@article{adolfi2023successes,
  title={Successes and critical failures of neural networks in capturing human-like speech recognition},
  author={Adolfi, Federico and Bowers, Jeffrey S and Poeppel, David},
  journal={Neural Networks},
  volume={162},
  pages={199--211},
  year={2023},
  publisher={Elsevier}
}

@article{zhang2016comprehensive,
  title={A comprehensive evaluation of random vector functional link networks},
  author={Zhang, Le and Suganthan, Ponnuthurai N},
  journal={Information sciences},
  volume={367},
  pages={1094--1105},
  year={2016},
  publisher={Elsevier}
}

@inproceedings{Devlin2019BERTPO,
  title={{BERT}: Pre-training of Deep Bidirectional Transformers for Language Understanding},
  author={Jacob Devlin and Ming-Wei Chang and Kenton Lee and Kristina Toutanova},
  booktitle={North American Chapter of the Association for Computational Linguistics},
  year={2019},
}

@article{smiti2020critical,
  title={A critical overview of outlier detection methods},
  author={Smiti, Abir},
  journal={Computer Science Review},
  volume={38},
  pages={100306},
  year={2020},
  publisher={Elsevier}
}

@article{wang2022review,
  title={A review on extreme learning machine},
  author={Wang, Jian and Lu, Siyuan and Wang, Shui-Hua and Zhang, Yu-Dong},
  journal={Multimedia Tools and Applications},
  volume={81},
  number={29},
  pages={41611--41660},
  year={2022},
  publisher={Springer}
}

@article{cao2018review,
  title={A review on neural networks with random weights},
  author={Cao, Weipeng and Wang, Xizhao and Ming, Zhong and Gao, Jinzhu},
  journal={Neurocomputing},
  volume={275},
  pages={278--287},
  year={2018},
  publisher={Elsevier}
}

@article{zhang2023tsk,
  title={TSK fuzzy system fusion at sensitivity-ensemble-level for imbalanced data classification},
  author={Zhang, Yuanpeng and Wang, Guanjin and Huang, Xiuyu and Ding, Weiping},
  journal={Information Fusion},
  volume={92},
  pages={350--362},
  year={2023},
  publisher={Elsevier}
}

@article{yang2024session,
  title={A session-incremental broad learning system for motor imagery EEG classification},
  author={Yang, Yufei and Li, Mingai and Liu, Hanlin and Li, Zhi},
  journal={Biomedical Signal Processing and Control},
  volume={97},
  pages={106717},
  year={2024},
  publisher={Elsevier}
}

@article{chen2017broad,
  author = {Chen, C. L. P. and Liu, Z.},
  title = {Broad learning system: An effective and efficient incremental learning system without the need for deep architecture},
  journal = {IEEE Trans. Neural Netw. Learn. Syst.},
  volume = {29},
  number = {1},
  pages = {10--24},
  year = {2017}
}

@article{saputra2024blsf,
  title={{BLSF}: Adaptive Learning for Small-Sample Medical Data With Broad Learning System Forest Integration},
  author={Saputra, Dimas Chaerul Ekty and Sunat, Khamron and Ratnaningsih, Tri},
  journal={IEEE Access},
  year={2024},
  publisher={IEEE}
}

@article{chen2018universal,
  title={Universal approximation capability of broad learning system and its structural variations},
  author={Chen, CL Philip and Liu, Zhulin and Feng, Shuang},
  journal={IEEE Transactions on Neural Networks and Learning Systems},
  volume={30},
  number={4},
  pages={1191--1204},
  year={2018},
  publisher={IEEE}
}

@article{feng2018broad,
  author = {Feng, S. and Chen, C. L. P.},
  title = {Broad learning system for control of nonlinear dynamic systems},
  journal = {Proc. IEEE Int. Conf. Syst., Man, Cybern. (SMC)},
  pages = {2230--2235},
  year = {2018}
}

@article{wang2021multi,
  title={Multi-modal broad learning for material recognition},
  author={Wang, Zhaoxin and Liu, Huaping and Xu, Xinying and Sun, Fuchun},
  journal={Cognitive Computation and Systems},
  volume={3},
  number={2},
  pages={123--130},
  year={2021},
  publisher={Wiley Online Library}
}

@article{gallicchio2020deep,
  author = {Gallicchio, C. and Scardapane, S.},
  title = {Deep Randomized Neural Networks},
  journal = {Recent Trends Learn. From Data},
  pages = {43--68},
  year = {2020}
}

@article{chu2024efficient,
  title={Efficient and effective ensemble broad learning system based on structural diversity},
  author={Chu, Fei and Wang, Jianwen and Cao, Yiwan and Li, Shuai},
  journal={Applied Soft Computing},
  volume={167},
  pages={112412},
  year={2024},
  publisher={Elsevier}
}

@article{huang2006extreme,
  title={Extreme learning machine: theory and applications},
  author={Huang, Guang-Bin and Zhu, Qin-Yu and Siew, Chee-Kheong},
  journal={Neurocomputing},
  volume={70},
  number={1-3},
  pages={489--501},
  year={2006},
  publisher={Elsevier}
}

@article{feng2018fuzzy,
  title={Fuzzy broad learning system: A novel neuro-fuzzy model for regression and classification},
  author={Feng, Shuang and Chen, CL Philip},
  journal={IEEE transactions on cybernetics},
  volume={50},
  number={2},
  pages={414--424},
  year={2018},
  publisher={IEEE}
}

@ARTICLE{10416391,
  author={Sajid, M. and Malik, A. K. and Tanveer, M. and Suganthan, Ponnuthurai N.},
  journal={IEEE Transactions on Fuzzy Systems}, 
  title={Neuro-Fuzzy Random Vector Functional Link Neural Network for Classification and Regression Problems}, 
  year={2024},
  volume={32},
  number={5},
  pages={2738-2749},
  doi={10.1109/TFUZZ.2024.3359652}}

@article{suganthan2021origins,
  author = {Suganthan, P. N. and Katuwal, R.},
  title = {On the origins of randomization-based feedforward neural networks},
  journal = {Appl. Soft Comput.},
  volume = {105},
  pages = {107239},
  year = {2021}
}

@article{lampert2013attribute,
  title={Attribute-based classification for zero-shot visual object categorization},
  author={Lampert, Christoph H and Nickisch, Hannes and Harmeling, Stefan},
  journal={IEEE Transactions on Pattern Analysis and Machine Intelligence},
  volume={36},
  number={3},
  pages={453--465},
  year={2013},
  publisher={IEEE}
}

@article{wang2023broad,
  title={Broad learning system with Takagi--Sugeno fuzzy subsystem for tobacco origin identification based on near infrared spectroscopy},
  author={Wang, Di and Yang, Simon X},
  journal={Applied Soft Computing},
  volume={134},
  pages={109970},
  year={2023},
  publisher={Elsevier}
}

@article{guo2020efficient,
  title={An efficient model for predicting setting time of cement based on broad learning system},
  author={Guo, Jifeng and Wang, Lin and Fan, Kaipeng and Yang, Bo},
  journal={Applied Soft Computing},
  volume={96},
  pages={106698},
  year={2020},
  publisher={Elsevier}
}

@article{su2023multi,
  title={Multi-Attn BLS: Multi-head attention mechanism with broad learning system for chaotic time series prediction},
  author={Su, Liyun and Xiong, Lang and Yang, Jialing},
  journal={Applied Soft Computing},
  volume={132},
  pages={109831},
  year={2023},
  publisher={Elsevier}
}

@article{yang2021broad,
  title={Broad learning extreme learning machine for forecasting and eliminating tremors in teleoperation},
  author={Yang, Qiye and Liang, Ke and Su, Tiecheng and Geng, Kuihua and Pan, Mingzhang},
  journal={Applied Soft Computing},
  volume={112},
  pages={107863},
  year={2021},
  publisher={Elsevier}
}

@article{malik2023random,
  title={Random vector functional link network: Recent developments, applications, and future directions},
  author={Malik, Ashwani Kumar and Gao, Ruobin and Ganaie, MA and Tanveer, Muhammad and Suganthan, Ponnuthurai Nagaratnam},
  journal={Applied Soft Computing},
  volume={143},
  pages={110377},
  year={2023},
  publisher={Elsevier}
}

@article{gong2021research,
  title={Research review for broad learning system: Algorithms, theory, and applications},
  author={Gong, Xinrong and Zhang, Tong and Chen, CL Philip and Liu, Zhulin},
  journal={IEEE Transactions on Cybernetics},
  volume={52},
  number={9},
  pages={8922--8950},
  year={2021},
  publisher={IEEE}
}

@article{deng2025robust,
  title={Robust dual-model collaborative broad learning system for classification under label noise environments},
  author={Deng, Wu and Shen, Jiuru and Ding, Jianming and Zhao, Huimin},
  journal={IEEE Internet of Things Journal},
  year={2025},
  publisher={IEEE}
}

@article{igelnik1995stochastic,
  author = {Igelnik, B. and Pao, Y.-H.},
  title = {Stochastic choice of basis functions in adaptive function approximation and the functional-link net},
  journal = {IEEE Trans. Neural Netw.},
  volume = {6},
  number = {6},
  pages = {1320--1329},
  year = {1995}
}

@article{xu2017re,
  title={Re-weighted discriminatively embedded $ k $-means for multi-view clustering},
  author={Xu, Jinglin and Han, Junwei and Nie, Feiping and Li, Xuelong},
  journal={IEEE Transactions on Image Processing},
  volume={26},
  number={6},
  pages={3016--3027},
  year={2017},
  publisher={IEEE}
}

@article{chen2010predictive,
  author = {Chen, Ning and Zhu, Jun and Xing, Eric},
  journal = {Advances in Neural Information Processing Systems},
  title = {Predictive subspace learning for multi-view data: a large margin approach},
  volume = {23},
  year = {2010}
}

@article{farquhar2005two,
  author = {Farquhar, Jason and Hardoon, David and Meng, Hongying and Shawe-Taylor, John and Szedmak, Sandor},
  journal = {Advances in Neural Information Processing Systems},
  title = {Two view learning: {SVM-2K}, theory and practice},
  volume = {18},
  year = {2005}
}

@article{hu2024multiview,
  author = {Hu, Kun and Xiao, Yingyuan and Zheng, Wenguang and Zhu, Wenxin and Hsu, Ching-Hsien},
  journal = {IEEE Transactions on Neural Information Processing Systems},
  title = {Multiview large margin distribution machine},
  year = {2024},
  publisher = {IEEE}
}

@article{li2016low,
  author = {Li, Jingjing and Wu, Yue and Zhao, Jidong and Lu, Ke},
  journal = {IEEE Transactions on Cybernetics},
  title = {Low-rank discriminant embedding for multiview learning},
  volume = {47},
  number = {11},
  pages = {3516--3529},
  year = {2016},
  publisher = {IEEE}
}

@article{li2018survey,
  author = {Li, Yingming and Yang, Ming and Zhang, Zhongfei},
  journal = {IEEE Transactions on Knowledge and Data Engineering},
  title = {A survey of multi-view representation learning},
  volume = {31},
  number = {10},
  pages = {1863--1883},
  year = {2018},
  publisher = {IEEE}
}

@article{quadir2024multiview,
  author = {Quadir, Abdul and Sajid, M and Tanveer, M},
  journal = {arXiv preprint arXiv:2409.02588},
  title = {Multiview random vector functional link network for predicting {DNA}-binding proteins},
  year = {2024}
}

@article{shi2020multi,
  author = {Shi, Zhenhua and Chen, Xiaomo and Zhao, Changming and He, He and Stuphorn, Veit and Wu, Dongrui},
  journal = {IEEE Transactions on Neural Systems and Rehabilitation Engineering},
  title = {Multi-view broad learning system for primate oculomotor decision decoding},
  volume = {28},
  number = {9},
  pages = {1908--1920},
  year = {2020},
  publisher = {IEEE}
}

@article{tang2020cgd,
  author = {Tang, Chang and Liu, Xinwang and Zhu, Xinzhong and Zhu, En and Luo, Zhigang and Wang, Lizhe and Gao, Wen},
  booktitle = {Proceedings of the AAAI Conference on Artificial Intelligence},
  title = {{CGD}: Multi-view clustering via cross-view graph diffusion},
  volume = {34},
  number = {04},
  pages = {5924--5931},
  year = {2020}
}

@article{tanveer2025grvfl,
  author = {Tanveer, M and Sharma, RK and Sajid, M and Quadir, A},
  journal = {Information Sciences},
  title = {{GRVFL-MV}: Graph random vector functional link based on multi-view learning},
  volume = {704},
  pages = {121947},
  year = {2025},
  publisher = {Elsevier}
}

@article{xie2015multi,
  author = {Xie, Xijiong and Sun, Shiliang},
  journal = {Intelligent Data Analysis},
  title = {Multi-view twin support vector machines},
  volume = {19},
  number = {4},
  pages = {701--712},
  year = {2015},
  publisher = {SAGE Publications Sage UK: London, England}
}

@article{xie2023deep,
  author = {Xie, Xijiong and Li, Yanfeng and Sun, Shiliang},
  journal = {Information Fusion},
  title = {Deep multi-view multiclass twin support vector machines},
  volume = {91},
  pages = {80--92},
  year = {2023},
  publisher = {Elsevier}
}

@article{yao2017learning,
  author = {Yao, Tingting and Wang, Zhiyong and Xie, Zhao and Gao, Jun and Feng, David Dagan},
  journal = {Pattern Recognition},
  title = {Learning universal multiview dictionary for human action recognition},
  volume = {64},
  pages = {236--244},
  year = {2017},
  publisher = {Elsevier}
}

@article{yu2025review,
  author = {Yu, Zhiwen and Dong, Ziyang and Yu, Chenchen and Yang, Kaixiang and Fan, Ziwei and Chen, CL Philip},
  journal = {Frontiers of Computer Science},
  title = {A review on multi-view learning},
  volume = {19},
  number = {7},
  pages = {197334},
  year = {2025},
  publisher = {Springer}
}

@article{gohain2023distance,
  title={A distance measure for optimistic viewpoint of the information in interval-valued intuitionistic fuzzy sets and its applications},
  author={Gohain, Brindaban and Chutia, Rituparna and Dutta, Palash},
  journal={Engineering Applications of Artificial Intelligence},
  volume={119},
  pages={105747},
  year={2023},
  publisher={Elsevier}
}

@incollection{atanassov1999intuitionistic,
  title={Intuitionistic fuzzy sets},
  author={Atanassov, Krassimir T},
  booktitle={Intuitionistic fuzzy sets: theory and applications},
  pages={1--137},
  year={1999},
  publisher={Springer}
}

@article{demvsar2006statistical,
  title={Statistical comparisons of classifiers over multiple data sets},
  author={Dem{\v{s}}ar, Janez},
  journal={The Journal of Machine Learning Research},
  volume={7},
  pages={1--30},
  year={2006},
  publisher={JMLR. org}
}

@article{haris2024breaking,
  title={Breaking down multi-view clustering: A comprehensive review of multi-view approaches for complex data structures},
  author={Haris, Muhammad and Yusoff, Yusliza and Zain, Azlan Mohd and Khattak, Abid Saeed and Hussain, Syed Fawad},
  journal={Engineering Applications of Artificial Intelligence},
  volume={132},
  pages={107857},
  year={2024},
  publisher={Elsevier}
}

@article{subramani2025improving,
  title={Improving Deep Random Vector Functional Link Networks through computational optimization of regularization parameters},
  author={Subramani, Chinnamuthu and Jagannath, Ravi Prasad K and Kuppili, Venkatanareshbabu},
  journal={Engineering Applications of Artificial Intelligence},
  volume={148},
  pages={110389},
  year={2025},
  publisher={Elsevier}
}

@article{huang2023gfbls,
  title={GFBLS: Graph-regularized fuzzy broad learning system for detection of interictal epileptic discharges},
  author={Huang, Zixuan and Duan, Junwei},
  journal={Engineering Applications of Artificial Intelligence},
  volume={125},
  pages={106763},
  year={2023},
  publisher={Elsevier}
}

@article{friedman1940comparison,
  title={A comparison of alternative tests of significance for the problem of m rankings},
  author={Friedman, Milton},
  journal={The Annals of Mathematical Statistics},
  volume={11},
  number={1},
  pages={86--92},
  year={1940},
  publisher={JSTOR}
}

@article{derrac2015keel,
  title={Keel data-mining software tool: Data set repository, integration of algorithms and experimental analysis framework},
  author={Derrac, J and Garcia, S and Sanchez, L and Herrera, F},
  journal={J. Mult. Valued Logic Soft Comput},
  volume={17},
  pages={255--287},
  year={2015}
}

@article{bergstra2012random,
  title={Random search for hyper-parameter optimization},
  author={Bergstra, James and Bengio, Yoshua},
  journal={The Journal of Machine Learning Research},
  volume={13},
  number={1},
  pages={281--305},
  year={2012},
  publisher={JMLR. org}
}

@misc{asuncion2007uci,
  title={{UCI} machine learning repository},
  author={Asuncion, Arthur and Newman, David and others},
  year={2007},
  publisher={Irvine, CA, USA}
}

@article{sajid2024intuitionistic,
  title={Intuitionistic fuzzy broad learning system: Enhancing robustness against noise and outliers},
  author={Sajid, M and Malik, Ashwani Kumar and Tanveer, Muhammad},
  journal={IEEE Transactions on Fuzzy Systems},
  volume={32},
  number={8},
  pages={4460--4469},
  year={2024},
  publisher={IEEE}
}

@article{sugiyama2007dimensionality,
  title={Dimensionality reduction of multimodal labeled data by local fisher discriminant analysis.},
  author={Sugiyama, Masashi},
  journal={Journal of Machine Learning Research},
  volume={8},
  number={5},
  year={2007}
}

@article{arora2026robust,
  title={A robust multi-view support vector machine with the RoBoSS loss function},
  author={Arora, Yash and Gupta, SK and Tanveer, M},
  journal={Neural Networks},
  pages={108937},
  year={2026},
  publisher={Elsevier}
}

@article{nie2020unsupervised,
  title={Unsupervised and semisupervised projection with graph optimization},
  author={Nie, Feiping and Dong, Xia and Li, Xuelong},
  journal={IEEE Transactions on Neural Networks and Learning Systems},
  volume={32},
  number={4},
  pages={1547--1559},
  year={2020},
  publisher={IEEE}
}

@article{guo2024intuitionistic,
  title={Intuitionistic fuzzy stochastic configuration networks for solving binary classification problems},
  author={Guo, Lili and Zhu, Jianglan and Zhang, Chenglong and Ding, Shifei},
  journal={IEEE Transactions on Fuzzy Systems},
  volume={32},
  number={8},
  pages={4210--4219},
  year={2024},
  publisher={IEEE}
}

@article{zhao2017multi,
  author = {Zhao, Jing and Xie, Xijiong and Xu, Xin and Sun, Shiliang},
  journal = {Information Fusion},
  title = {Multi-view learning overview: Recent progress and new challenges},
  volume = {38},
  pages = {43--54},
  year = {2017},
  publisher = {Elsevier}
}

@article{yan2006graph,
  title={Graph embedding and extensions: A general framework for dimensionality reduction},
  author={Yan, Shuicheng and Xu, Dong and Zhang, Benyu and Zhang, Hong-Jiang and Yang, Qiang and Lin, Stephen},
  journal={IEEE Transactions on Pattern Analysis and Machine Intelligence},
  volume={29},
  number={1},
  pages={40--51},
  year={2006},
  publisher={IEEE}
}

@article{houthuys2018multi,
  title={Multi-view least squares support vector machines classification},
  author={Houthuys, Lynn and Langone, Rocco and Suykens, Johan AK},
  journal={Neurocomputing},
  volume={282},
  pages={78--88},
  year={2018},
  publisher={Elsevier}
}

@article{huang2024stacking,
  author = {Huang, T. and others},
  title = {Stacking multi-view broad learning system with residual structures for classification},
  journal = {Inf. Sci.},
  volume = {669},
  pages = {120559},
  year = {2024}
}

@article{shi2023graph,
  title={Graph embedding deep broad learning system for data imbalance fault diagnosis of rotating machinery},
  author={Shi, Mingkuan and Ding, Chuancang and Wang, Rui and Shen, Changqing and Huang, Weiguo and Zhu, Zhongkui},
  journal={Reliability Engineering \& System Safety},
  volume={240},
  pages={109601},
  year={2023},
  publisher={Elsevier}
}

@article{liu2021graph,
  title={Graph-based broad learning system for classification},
  author={Liu, Zheng and Huang, Shiluo and Jin, Wei and Mu, Ying},
  journal={Neurocomputing},
  volume={463},
  pages={535--544},
  year={2021},
  publisher={Elsevier}
}

@article{li2023local,
  title={Local discriminative embedding broad learning system with graph convolutional for hyperspectral image classification},
  author={Li, Wei and Shi, Yuanquan and Li, Liyun and Ma, Xiangbo},
  journal={IEEE Access},
  volume={11},
  pages={91879--91890},
  year={2023},
  publisher={IEEE}
}

@inproceedings{jiang2019fast,
  title={A Fast Approach of Graph Embedding Using Broad Learning System},
  author={Jiang, Long and Zuo, Yi and Li, Tieshan and Chen, CL Philip},
  booktitle={International Conference on Multidisciplinary Social Networks Research},
  pages={164--172},
  year={2019},
  organization={Springer}
}

@article{malik2022alzheimer,
  title={Alzheimer’s disease diagnosis via intuitionistic fuzzy random vector functional link network},
  author={Malik, Ashwani Kumar and Ganaie, MA and Tanveer, M and Suganthan, PN},
  journal={IEEE Transactions on Computational Social Systems},
  volume={11},
  number={4},
  pages={4754--4765},
  year={2022},
  publisher={IEEE}
}

@article{chen2025adaptive,
  title={Adaptive broad network with graph-fuzzy embedding for imbalanced noise data},
  author={Chen, Wuxing and Yang, Kaixiang and Yu, Zhiwen and Nie, Feiping and Chen, CL Philip},
  journal={IEEE Transactions on Fuzzy Systems},
  year={2025},
  publisher={IEEE}
}

@article{gao2024interactive,
  title={An interactive approach for intuitionistic fuzzy data envelopment analysis},
  author={Gao, Qingping and Lin, Yu and Luo, Yu and Xu, Xuewen},
  journal={IEEE Transactions on Fuzzy Systems},
  volume={32},
  number={9},
  pages={5189--5200},
  year={2024},
  publisher={IEEE}
}

\end{document}